%% file: main.tex
\documentclass[10pt,twocolumn,letterpaper]{article}

\usepackage{iccv}
\usepackage{times}
\usepackage{epsfig}
\usepackage{graphicx}
\usepackage{amsmath}
\usepackage{amssymb}
\usepackage[export]{adjustbox}
\usepackage{enumitem}
\usepackage{subcaption} 
\usepackage{wrapfig}
\usepackage{rotating}
\usepackage{tabularx,booktabs} 

\usepackage[pagebackref=true,breaklinks=true,letterpaper=true,colorlinks,bookmarks=false]{hyperref}

\newcommand{\comment}[1]{}
\newcommand{\norm}[1]{\left\lVert#1\right\rVert}
\newcommand{\todo}{\textbf{\color{red}{TODO:}}}
\newcommand{\Lrecon}{\mathcal{L}_\texttt{reconst}}

\iccvfinalcopy 

\def\iccvPaperID{1017} 

\ificcvfinal\fi
\begin{document}

\title{InGAN: Capturing and Remapping the ``DNA" of a Natural Image}

\author{Assaf Shocher$^*$
\qquad
Shai Bagon$^\dagger$
\qquad
Phillip Isola$^\ddagger$
\qquad
Michal Irani$^*$
\\
\small 
$^*$Dept. of Computer Science and Applied Math,
The Weizmann Institute of Science\\
\small
$^\dagger$ Weizmann Center of Artificial Intelligence (WAIC)\\
\small
$^\ddagger$Computer Science and Artificial Intelligence Lab
Massachusetts Institute of Technology\\
\small
{\textbf{\emph{Project Website:}}}  \url{http://www.wisdom.weizmann.ac.il/\~vision/ingan/}}

\maketitle

\input{fig_all_sizes}

\input{abstract}

\input{fig_scheme}

\input{introduction}

\input{fig_g}
\input{method}

\input{implementation}

\input{fig_natural_image_remapping}

\input{unified-framework}

\input{experiments}

\input{acknowledge}

{\small
\bibliographystyle{ieee}
\bibliography{references}
}

\end{document}

%% file: fig_all_sizes.tex
\begin{figure*}
\centering
\includegraphics[width=1.98\columnwidth]{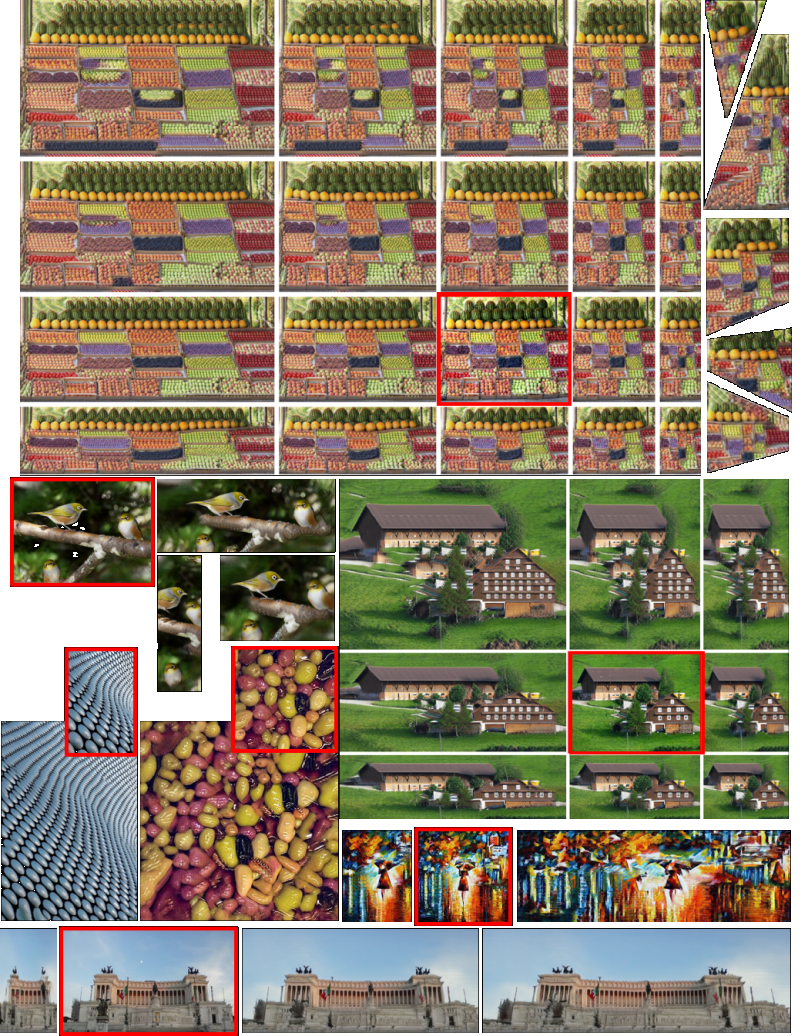}
\vspace*{-0.3 cm}
\caption{\label{fig:all_sizes}
\textbf{InGAN's Capabilities:} \ 
\small \it 
\emph{(Top:)} Once trained on an input image (marked by a {\color{red}{red frame}}), InGAN can synthesize a plethora of new  images of significantly different sizes/shapes/aspect-ratios – all with the same ``DNA'' of the input image. All elements inside the image maintain their {local size/shape} and relative position. \textbf{Please view attached videos to see the continuum between different shapes/sizes/aspect-ratios}.\  \ \ 
\emph{(Bottom:)} \ InGAN  provides a \emph{unified treatment for a variety of  different datatypes} -- single/multi-texture images, painitings, and complex natural images,  all under a single umbrella.
}
\end{figure*}

%% file: abstract.tex
\begin{abstract}

Generative Adversarial Networks (GANs) typically learn a \underline{distribution of images} in a large image dataset, and are then able to generate new images from this distribution. However, each natural  image has its own internal statistics, captured by its unique \underline{distribution of patches}. In this paper we propose an ``Internal GAN'' (InGAN) -- \textbf{an image-specific GAN} -- which trains on a single input image and learns its internal {distribution of patches}. It is then able to synthesize a plethora of new natural images of significantly different sizes, shapes and aspect-ratios – all with the same internal patch-distribution (same ``DNA'') as the input image. In particular, despite large changes in \underline{global size/shape} of the image, all elements inside the image maintain their \underline{local size/shape}. InGAN is fully unsupervised, requiring no additional data other than the input image itself. Once trained on the input image, it can remap the input to any size or shape in a single feedforward pass, while preserving the same internal patch distribution. \mbox{InGAN} provides a unified framework for a variety of tasks, bridging the gap between textures and natural images.\footnote{Code will be made publicly available.}

\end{abstract}

%% file: fig_scheme.tex
\begin{figure*}

\begin{minipage}{.63\textwidth}
\centering
\includegraphics[width=\columnwidth]{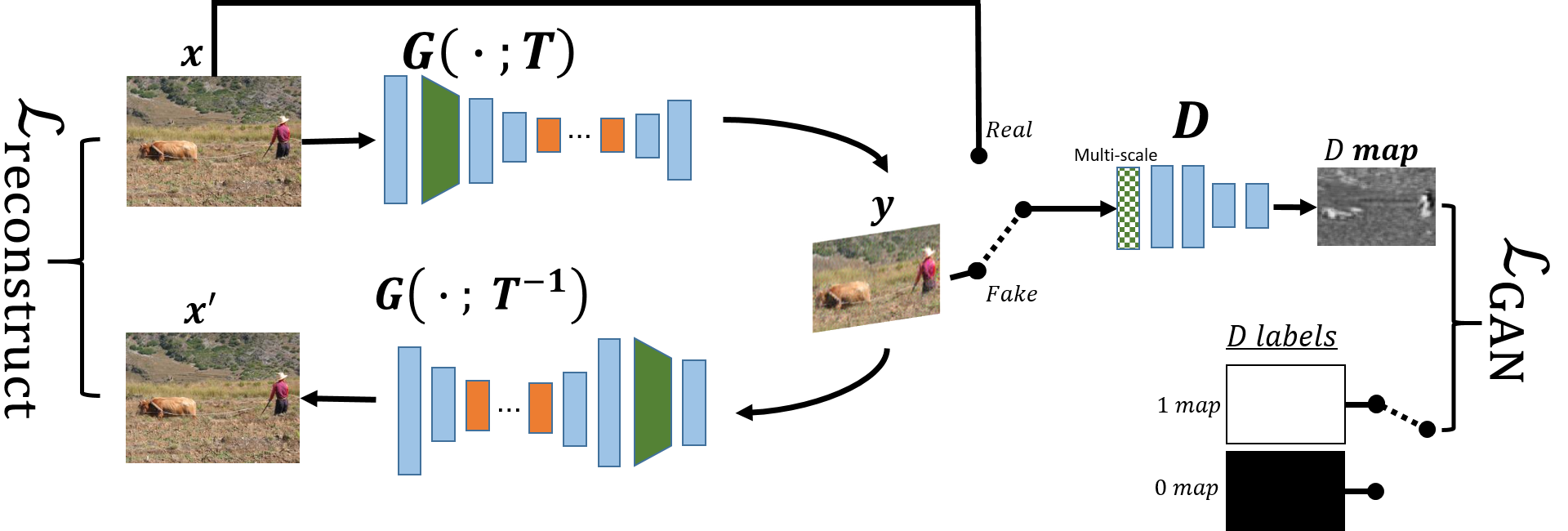}
\captionof{figure}{\label{fig:scheme}
\textbf{InGAN Architecture:} \ 
\small  \it
InGAN consists of a Generator $G$ that retargets input $x$ to output $y$ whose size/shape is determined by a geometric transformation $T$
(top left). A multiscale discriminator $D$ learns to discriminate the patch statistics of the fake output $y$ from the true patch statistics of the input image (right). Additionally, we take advantage of $G$'s automorphism to reconstruct the input back from $y$ using $G$ and the inverse transformation $T^{-1}$ (bottom left).
}
\end{minipage}
\hspace{1em}
\begin{minipage}{.37\textwidth}
\vspace*{-0.5cm}
\centering
\includegraphics[width=\columnwidth]{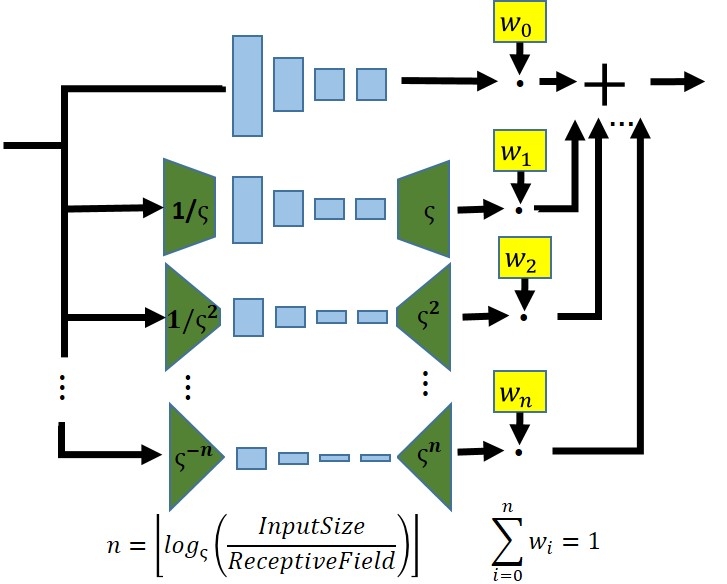}
\captionof{figure}{\label{fig:d}
\textbf{{Adaptive Multi-Scale Patch Discriminator}} 
}
\end{minipage}
\vspace*{-0.5cm}
\end{figure*}

%% file: introduction.tex
\section{Introduction}
\label{sec:introduction}

Each natural image has its unique internal statistics: small patches (e.g., 5x5, 7x7) recur abundantly inside a single natural image~\cite{glasner2009sr,zontak2011internal}. This patch recurrence was shown to form \emph{a strong image-specific prior} for solving many ill-posed vision tasks in an unsupervised way~\cite{NLM2005,Dabov2007,Elad2006,efros1999texture,glasner2009sr,simakov2008summarizing,pritch2009shift,Barnes09,Cho_TPAMI2010}.
In this paper we \emph{capture and visualize} this unique image-specific patch-distribution, and map it to new target images of different sizes and shapes 
-- all with the same internal patch distribution as the input image (which we loosely call ``same DNA'').

For example, imagine you are given an input image, and you wish to transform it to a new image, of drastically different shape, size and aspect ratio. But you don't want to distort any of its internal elements; you want to keep them all in their original size, shape, aspect ratio, and in the same relative position within the image. Such examples are shown in Fig.~\ref{fig:all_sizes}. Note that despite changing the global size and shape of the farmhouse image, the windows in the target images maintain their local size and shape. Rows of windows are automatically added/removed, and likewise for the number of windows in each row. Similarly, when the fruit-stand image in Fig.~\ref{fig:all_sizes} is enlarged, more fruits are added in each fruit-box while keeping the size of each fruit the same; and vice versa – when the image grows smaller, the number of fruits grows smaller, while maintaining their size and their relative position within the image. Furthermore, note that the target image may not necessarily be rectangular.

How can this be done? One way to satisfy these criteria is to require that the distribution of patches in the target images match the distribution of patches in the input image, at multiple image scales. We propose \emph{Distribution-Matching} as a new objective for ``visual retargeting''. Note that we use  the term retargeting here differently than its common use in image-retargeting methods~\cite{seamcarving,cho2017weakly,wolf2007}. 
\emph{Distribution-matching} allows synthesizing new target images of different sizes and shapes –- all with the same internal patch distribution as the input image.

A closely related work 
is the \emph{Bidirectional-Similarity} of Simakov~et~al.~\cite{simakov2008summarizing}.
The Bidirectional objective constrains the target image to contain only patches from the input image (``Visual Coherence''), and vice versa,  the input should contain only patches from the target (``Visual Completeness''). Hence, no new artifacts are introduced in the target image and no critical information is lost either.
Our new ``\emph{Distribution Matching}'' formulation extends the Bidirectional-Similarity and goes beyond it in multiple ways: (i)~It requires not only that all input patches  be in the output (and vice versa), but also that the \emph{frequency} of these patches remain the same.
(ii)~By matching \emph{distributions} rather than individual patches, we can leverage recent advances in distribution modeling using Generative Adversarial Networks (GANs)~\cite{gans}. (iii)~A single forward pass through our trained network can generate target images of any size/shape/aspect ratio, without having to solve a new optimization problem for each  desired target.

GANs can be understood as a tool for distribution matching~\cite{gans}.
A GAN typically learns a \underline{distribution of images} in a large image dataset. It maps data sampled from one distribution to transformed data that is indistinguishable from a target distribution, \mbox{$G:x\rightarrow y$} with $x$$\sim$$p_{x}$, and $G(x)$$\sim$$p_{y}$.
We propose an ``Internal GAN'' (InGAN) -- \textbf{an image-specific GAN} -- which trains on a single input image and learns its unique internal \underline{distribution of patches}. InGAN is fully unsupervised, requiring no training examples other than the input image.
Unlike most GANs, which map between two different distributions, InGAN is an \emph{automorphism}, $G: x \rightarrow x$, with $p_x$ being the distribution of patches in the input image.  Retargeting is achieved by modifying the size and shape of the output tensor, which changes the arrangement of patches, but not the distribution of patches.

Although this formulation is sufficient in theory to encourage both \emph{Coherence} and \emph{Completeness}, in practice we observe that completeness is often not achieved -- many patches from the input image are omitted in the output (``mode collapse''). To ameliorate this, we introduce a second mechanism for encouraging completeness: it should be possible to reconstruct (``decode") the input image from the output, i.e. $\norm{F(G(x)) - x}$ should be small, where $F$ is a second network trained to perform the reverse mapping. This objective encourages the mapping between input and retargeted output to be {\emph cycle-consistent}~\cite{cyclegan}, a desideratum that has recently come into widespread use and often improves the results of distribution matching problems.
Since our proposed InGAN is an automorphism, we use $G$ itself to perform the decoding, that is $\norm{G\left(G\left(x\right)\right) - x}$ resulting in \emph{a novel \textbf{Encoder-Encoder} architecture}.

Our results reinforce the recent finding that neural nets, when trained on a \emph{single} image, can learn a useful representation of the \emph{internal statistics} of that image. These representations can then be used  to super-resolve the image~\cite{shocher2018zssr},  to inpaint patches removed from the image~\cite{ulyanov2018deep}, or to synthesize textures from the image~\cite{jetchev2016texture,nonstationary}. In particular, GANs trained on a single \emph{texture} image were introduced by~\cite{jetchev2016texture,nonstationary}.
Through InGAN, we further show that such \emph{image-specific} internal statistics, encoded in a feedforward neural net, provides a \emph{single unified framework for a variety of new tasks/capabilities} (Image-Retargeting, Image Summarization \& Expansion, Texture-Synthesis,  synthesizing \emph{Non-Rectangular} outputs, etc.) Through its multi-scale discriminator, InGAN further provides a \emph{unified treatment for a variety of  different datatypes} (single/multi-texture images, painitings, and complex natural images),  all under a single umbrella.
While not guaranteed to provide state-of-the-art results compared to  \emph{specialized} methods optimized for a specific  task/datatype, it compares favorably to them,
and further gives rise to \emph{new applications}.

\vspace*{0.2cm}
\noindent
Our contributions are several-fold:
\begin{itemize}[noitemsep,nolistsep,leftmargin=*]
\item We define \emph{distribution-matching of patches} as a criterion for visual retargeting and image manipulation.
\item InGAN provides a \emph{unified-framework} for various different tasks and different datatypes, all  with a single network architecture.
\item Once trained, InGAN can produce outputs of significantly different sizes, shapes, and aspect ratios, including \emph{non-rectangular} output images.
\item To the best of our knowedge, 
InGAN is the first to train a GAN on a single \emph{natural} image.
\item The inherent symmetry of the challenge (an Automorphism) gives rise to a \emph{new Encoder-Encoder architecture}.
\end{itemize}

%% file: fig_g.tex
\begin{figure}
\includegraphics[width=\columnwidth]{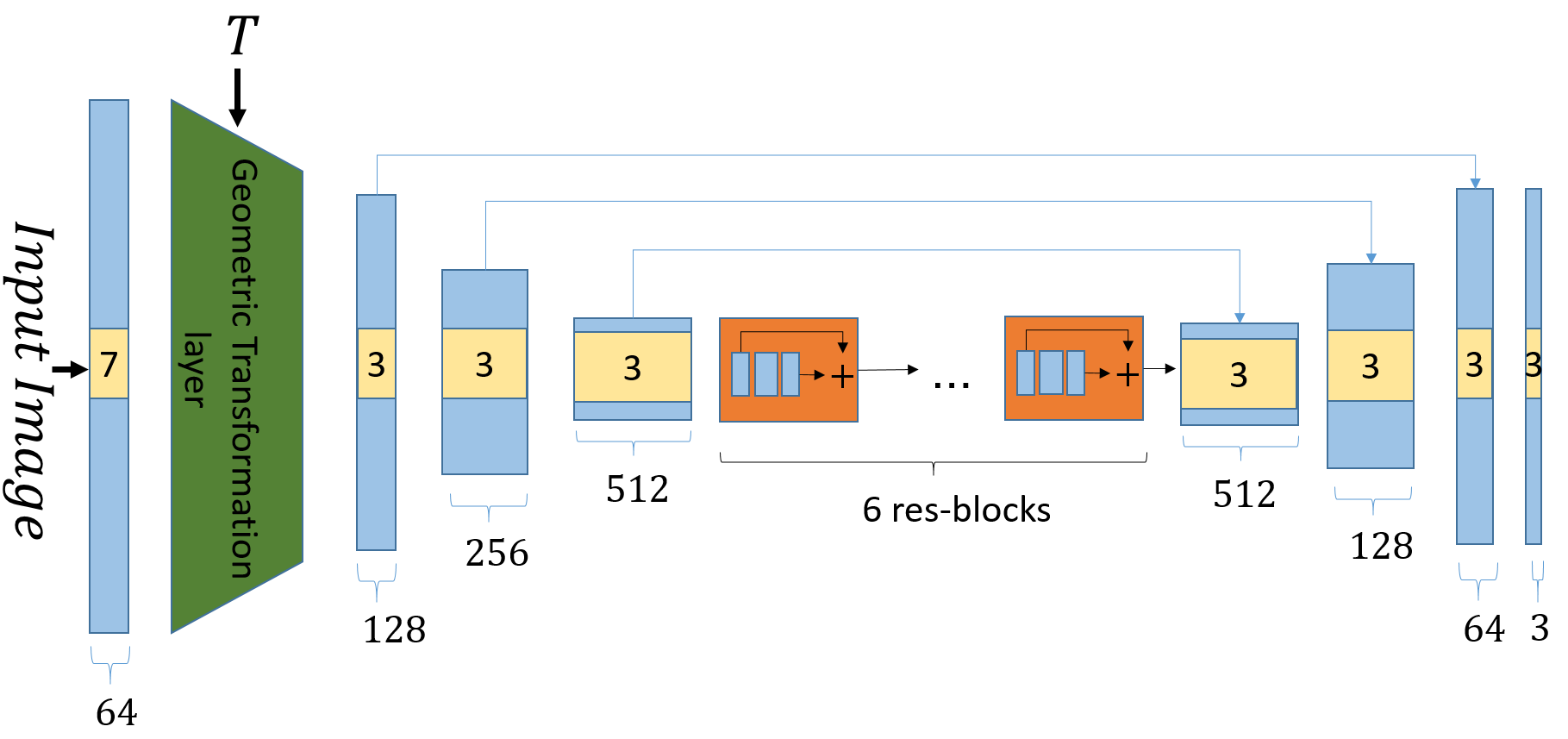}
\caption{\label{fig:g}
\textbf{Generator architecture:} 
\small \it $G$ receives an input image $x$ and a geometric transformation $T$ which determines the size/shape of the output.
}
\vspace*{-0.5cm}
\end{figure}

%% file: method.tex
\section{Method}
\subsection{Overview}

Our InGAN is an \emph{image-conditional} GAN (e.g., ~\cite{pix2pix2017}) that maps an input image (as opposed to noise) to a remapped target output image. It uses a generator, $G$, a discriminator, $D$, and re-uses $G$ for decoding/reconstructing the input given the output, as depicted in Fig.~\ref{fig:scheme}. Our formulation aims to achieve two properties:
(i)~\textbf{\mbox{matching distributions:}} The distribution of patches, across scales, in the synthesized image, should match that distribution in the original input image. This property is a generalization of both the \emph{Coherence} and \emph{Completeness} objectives of~\cite{simakov2008summarizing}. \ 
(ii)~\textbf{localization:} The elements' locations in the generated image should generally match their relative locations in the original input image.

In detail, our method works as follows. Given an input image $x$, and a geometric transformation $T$ (which determines the desired shape/size of the target output), $G$ synthesizes a new image, $y = G(x; T)$. 
For example, $T$ could be a scaling transformation, a skew, an affine transformation, or any other invertible geometric transformation. In our current implementation we allow $T$ to be any desired homography (a 2D projective transformation).
During training, $T$ is \emph{randomly sampled} at each iteration. Once trained, $G$ can handle any desired transformation $T$  (any homography).

The generator $G$ trains to output an image $y$ of  size/shape/aspect-ratio specified by $T$ that, at the patch level, is indistinguishable from the input image $x$, according to an adversarial discriminator $D$. We adopt the LSGAN~\cite{lsgan} variant of this optimization problem: $G^*$=$\min_G\max_D \mathcal{L}_{\texttt{GAN}}(G,D)$, where
$$\hspace*{-0.3cm}
\mathcal{L}_{\texttt{GAN}}(G,D) = \mathbb{E}_{y \sim p_{\text{data}}(x)}[(D(x)-1)^2]+\mathbb{E}_{x \sim p_{\text{data}}(x)}[D(G(x))^2]$$

The discriminator $D$ and $\mathcal{L}_\texttt{GAN}$ encourage matching the patch distribution of $y=G\left(x; T\right)$ to that of $x$. $D$ is fully convolutional:
it outputs a map (rather than a scalar) where each pixel depends only on its receptive field \cite{dumoulin2016guide},
thus it has all the \emph{patches} of the original input $x$ to train on.
Using a \emph{multiscale} $D$  enforces
patch distribution matching at each scale separately.

In practice using only $\mathcal{L}_\texttt{GAN}$ may result in mode collapse, i.e. the synthesized image consists of only a subset of patches of the original image (it is \emph{coherent}) but many patches are missing (it is not \emph{complete}). To ameliorate this mode collapse we take advantage of the automorphism of $G$ and re-use $G$ to reconstruct $x$ back from the synthesized image $y$. The $\ell_1$ reconstruction loss $\Lrecon = \norm{G\left(G\left(x; T\right); T^{-1}\right) - x}_1$ encourages $G$ to avoid mode collapse and maintain \emph{completeness}.
The overall loss function of InGAN is $\mathcal{L}_\texttt{InGAN}=\mathcal{L}_\texttt{GAN} + \lambda\cdot\Lrecon$

Localization is implicitly encouraged through the choice of network architecture. The architecture is locally-connected rather than fully-connected (in particular, it is convolutional). This means that an output pixel at location $\{i,j\}$ can only depend on input pixels in a finite receptive field around that location in the input image. Nonlocal mappings, beyond a certain radius, are impossible with such an architecture. We also conjecture that simple local mappings are easier to learn than nonlocal mappings, and convolutional networks may naturally converge to these solutions~\cite{galanti2017role}.
 
\input{fig_comparison_x2}

\subsection{Shape-flexible Generator} 
Fig.~\ref{fig:g} shows the architecture of the generator $G$.
The desired geometric transformation for the output shape $T$ is treated as an additional input that is fed to $G$ for every forward pass.
A \emph{parameter-free} transformation layer (green layer in Fig.~\ref{fig:g}) geometrically transforms the feature map to the desired output shape.
Making the transformation layer parameter-free allows training $G$ once to transform $x$ to any size, shape or aspect ratio at test time.

The generator is fully-convolutional with an hourglass architecture and skip connections (U-net~\cite{unet} architecture). 
The bottleneck consists of residual-blocks \cite{resnet}. 
Downscaling is done by max pooling. 
Upscaling is done by nearest-neighbor resizing followed by a convolutional layer~\cite{odena2016deconvolution}.

\subsection{Multi-scale Patch Discriminator}
\label{D} 
\input{fig_comparison_textures}
We use a fully-convolutional patch discriminator $D$ (Fig.~\ref{fig:d}), as introduced in \cite{pix2pix2017}. The labels for the discriminator are  \emph{maps} (matrices of real/fake labels) of same size as the desired output $y$. 
Thus $D$ grades each \emph{patch} for how well it matches the patch distribution, rather than grading the entire synthesized image. 

InGAN uses a multi-scale $D$ (similar to \cite{pix2pixHD}). This feature is significant: A single scale discriminator can only capture patch statistics of a specific size. Using a multiscale $D$ matches the patch distribution over a range of patch sizes, capturing both fine-grained details as well as coarse structures in the image. 
At each scale, the discriminator is rather simple: it consists of just four conv-layers with the first one strided.
Weights are \emph{not shared} between different scale discriminators. The downsampling factor from one scale to the next is set to $\varsigma = \sqrt{2}$.

The multiscale $D$ outputs $n$ discrimination maps that are summed via global weighted mean pooling to yield $D$'s output. The weights are updated during the optimization process in a coarse-to-fine manner. Initially, the weights are such that most of the contribution to $\mathcal{L}_\texttt{GAN}$ is from the coarsest scale. As the training progresses, the contribution gradually shifts to the finer scales.

\subsection{Generator Invertibillity}
\label{invg}
Training $G$ with $\mathcal{L}_\texttt{GAN}$ often leads to mode collapse where the synthesized $y$'s are  \emph{coherent}  --  the multiscale patches of $y$ are drawn from the input image's distribution -- but not \emph{complete} -- i.e. important visual information is missing from the generated $y$'s.
To achieve better completeness, \mbox{InGAN} reconstructs the input image $x$ from the output image $y$, ensuring no visual information was lost in $y$.
Taking advantage of $G$'s automorphism allows us to re-use $G$ to reconstruct $x$ back from $y$ without training an additional decoder, yielding an ``Encoder-Encoder" architecture.

%% file: fig_comparison_x2.tex
\begin{figure*}
\centering
\begin{tabular}{c@{\hskip1pt}|@{\hskip1pt}c@{\hskip1pt}|@{\hskip1pt}c@{\hskip1pt}|@{\hskip1pt}c@{\hskip1pt}|@{\hskip1pt}c@{\hskip1pt}|@{\hskip1pt}c}
     input & 
     Seam-Carving \cite{seamcarving-imp} & 
     BiDir \cite{simakov2008summarizing}  & 
     Spatial-GAN \cite{jetchev2016texture} &
     Non-stationary \cite{nonstationary} & 
     \textcolor{red}{\textbf{InGAN (Ours)}}\\

     \includegraphics[width=.09\linewidth, cframe=red 1.6pt]{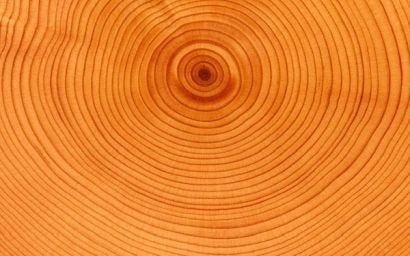} &
     \includegraphics[width=.18\linewidth]{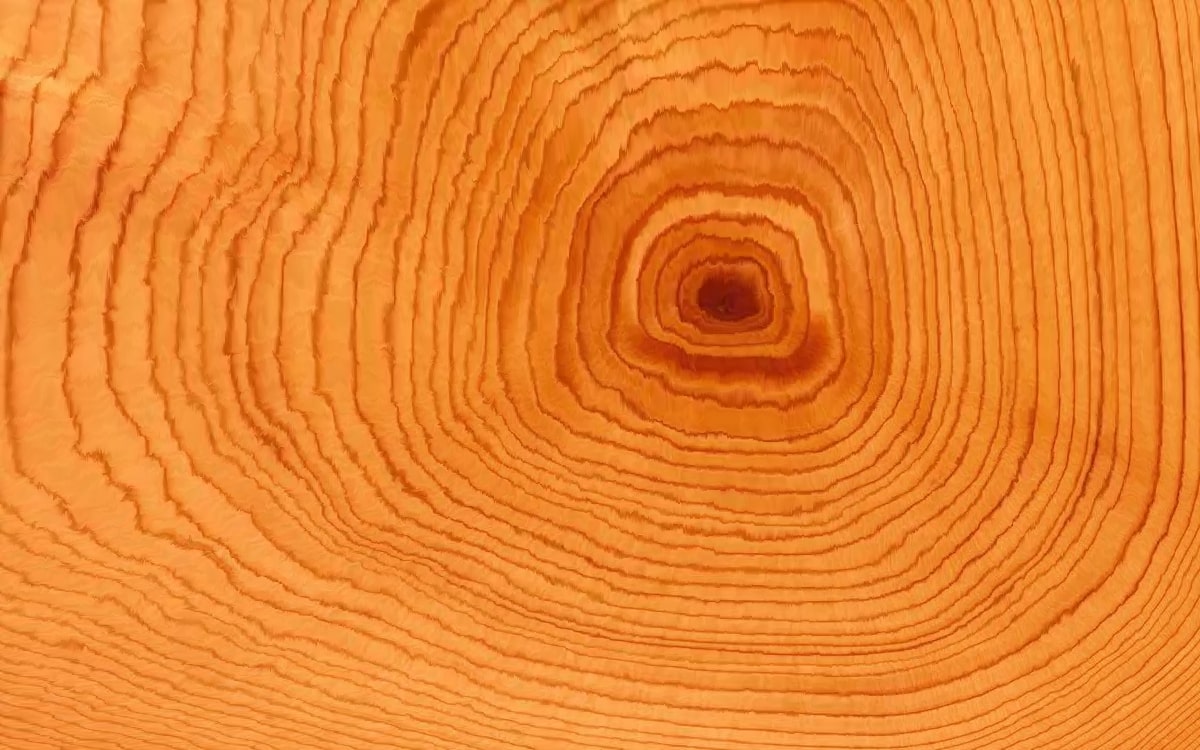} & 
     \includegraphics[width=.18\linewidth]{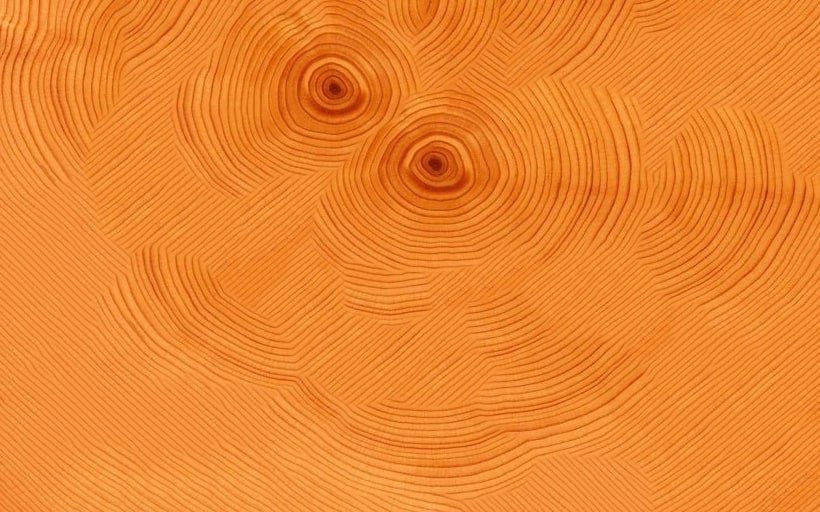} &
     \includegraphics[width=.18\linewidth]{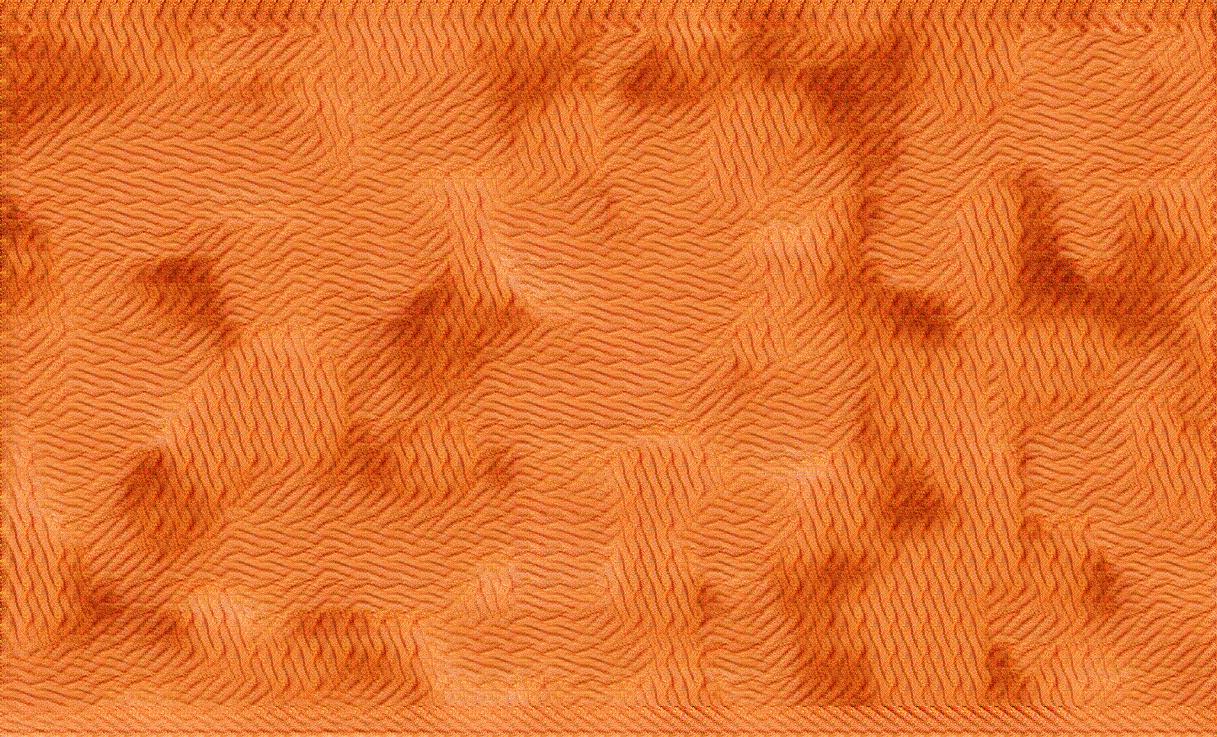} &
     \includegraphics[width=.18\linewidth]{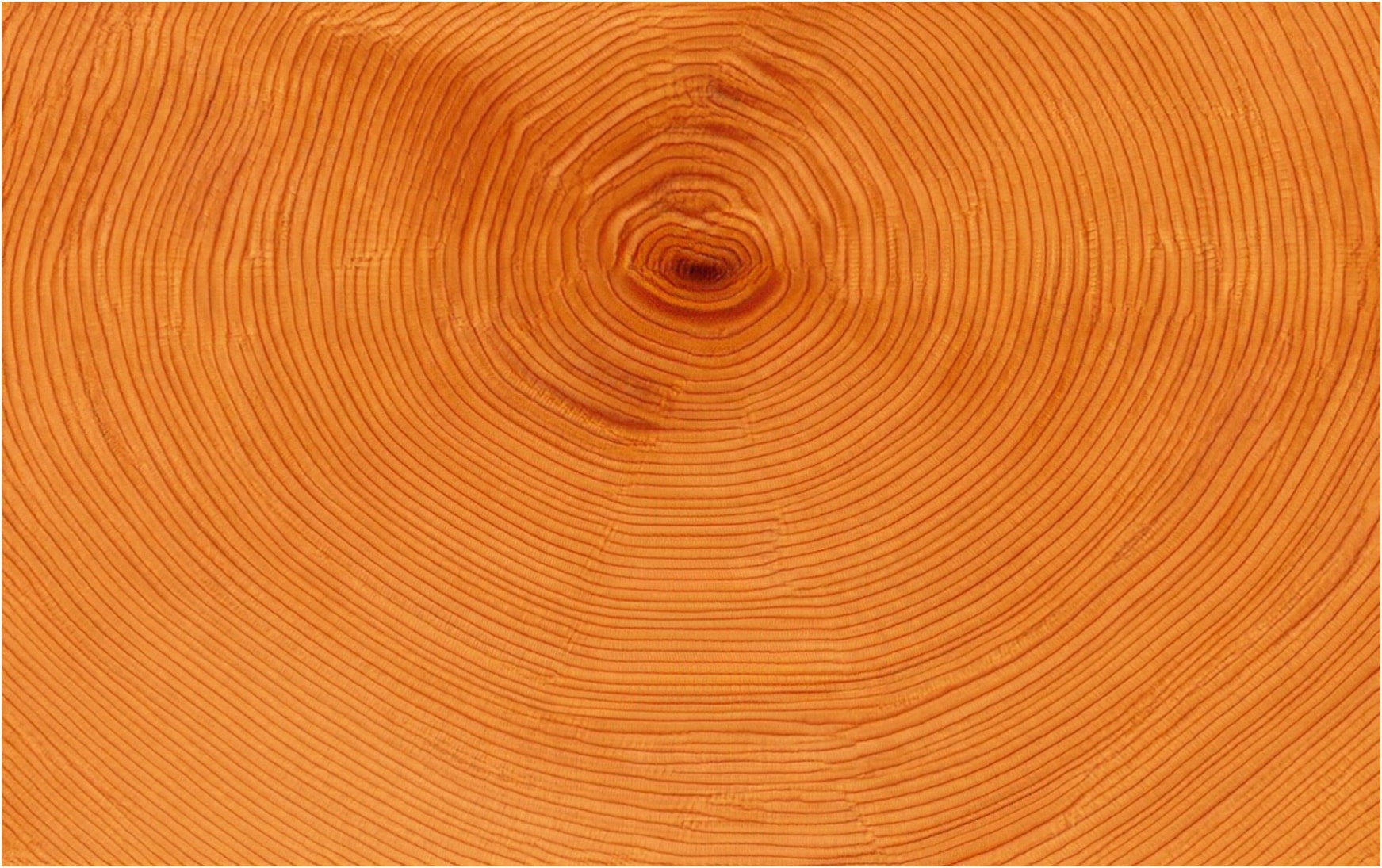} & 
     \includegraphics[width=.18\linewidth]{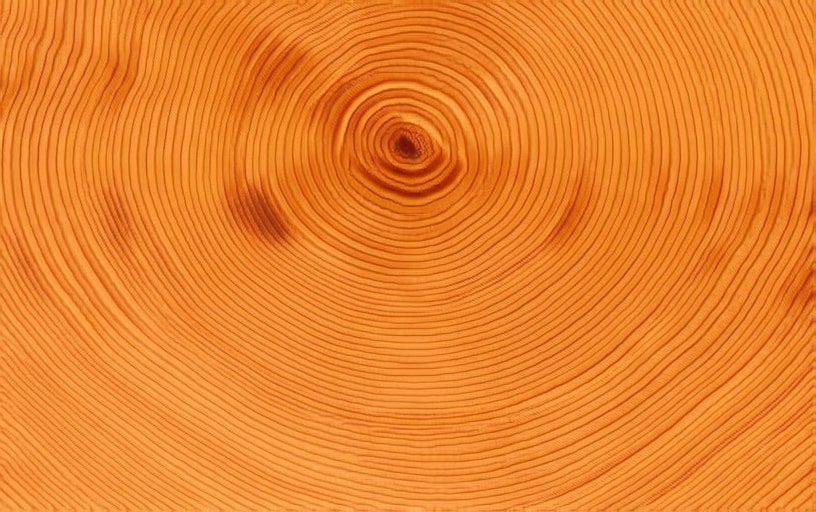} \\

     \includegraphics[width=.09\linewidth, cframe=red 1.6pt]{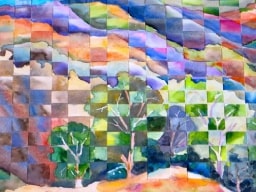} &
     \includegraphics[width=.18\linewidth]{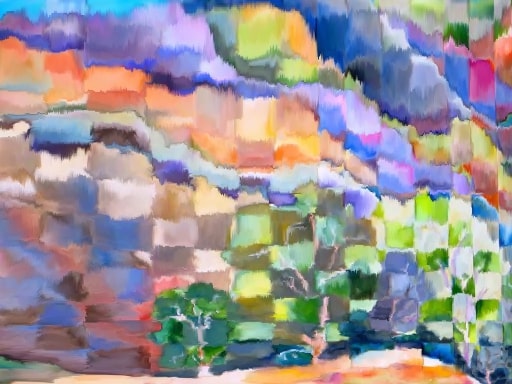} &
     \includegraphics[width=.18\linewidth]{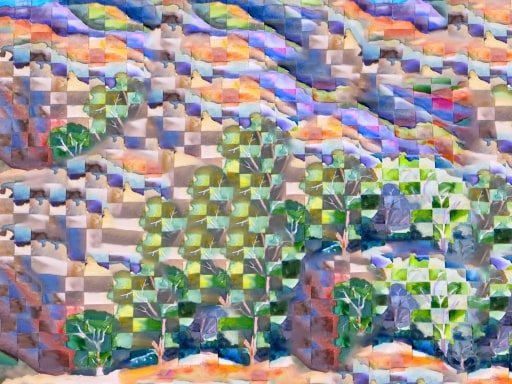} &
     \includegraphics[width=.18\linewidth]{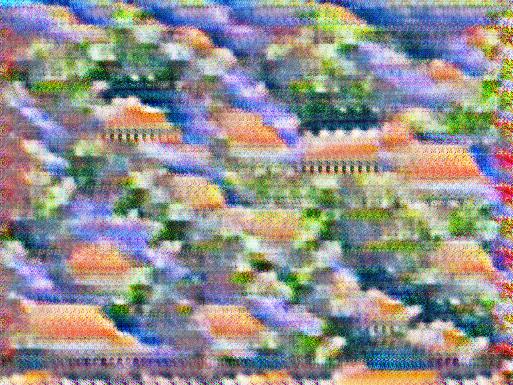} &
     \includegraphics[width=.18\linewidth]{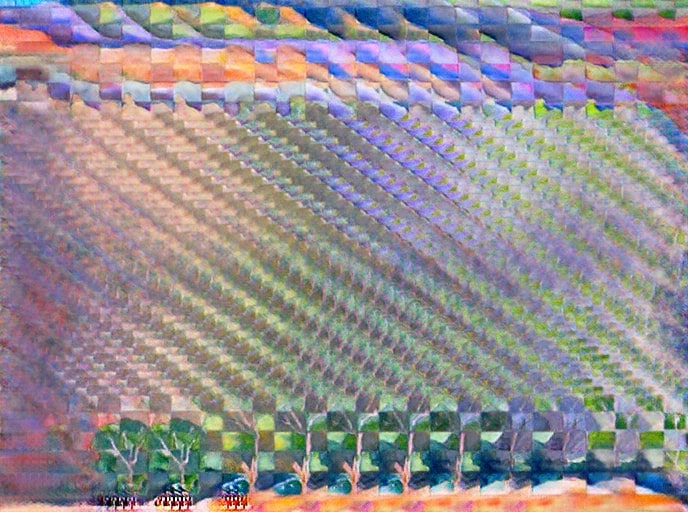} &
     \includegraphics[width=.18\linewidth]{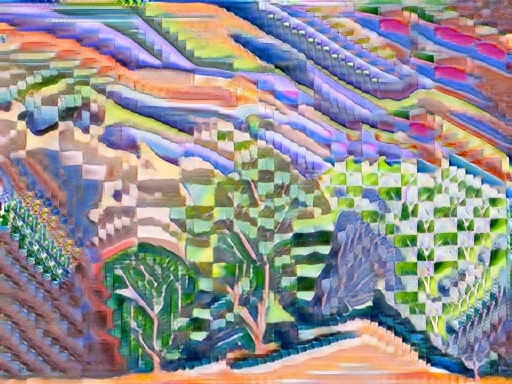}  \\
     
     \includegraphics[width=.09\linewidth, cframe=red 1.6pt]{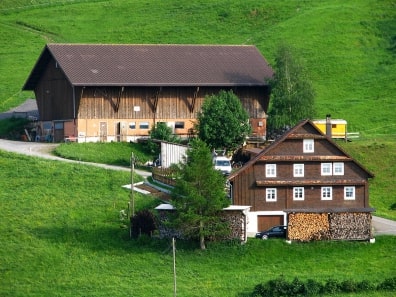} & 
     \includegraphics[width=.18\linewidth]{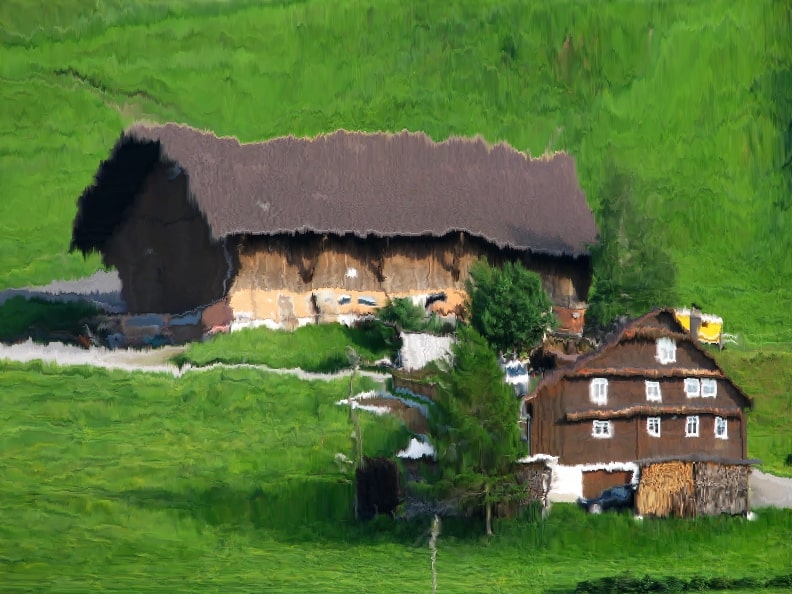} &
     \includegraphics[width=.18\linewidth]{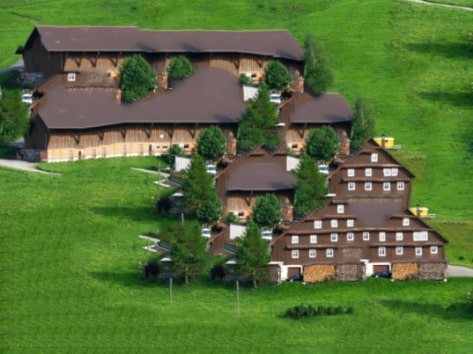} &
     \includegraphics[width=.18\linewidth]{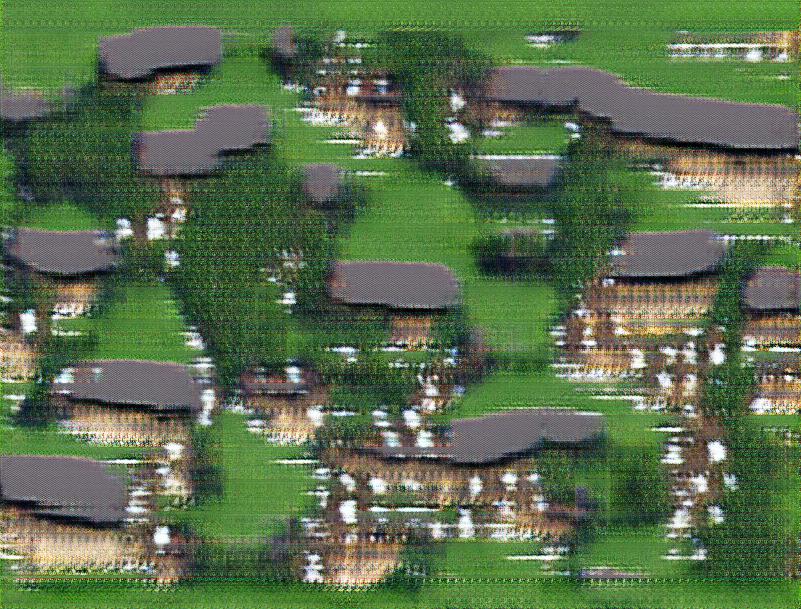} &
     \includegraphics[width=.18\linewidth]{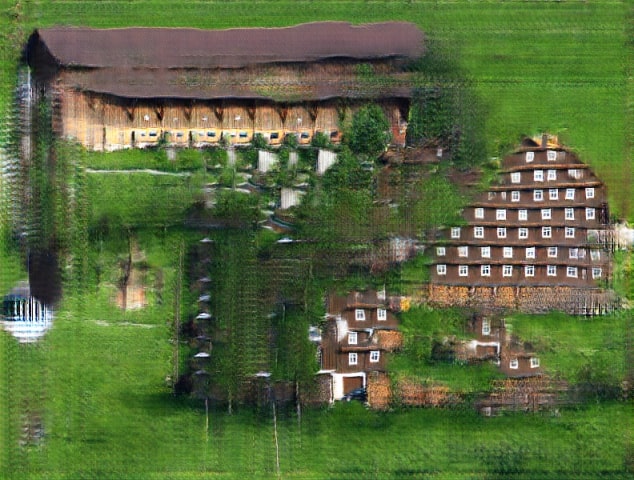} &
     \includegraphics[width=.18\linewidth]{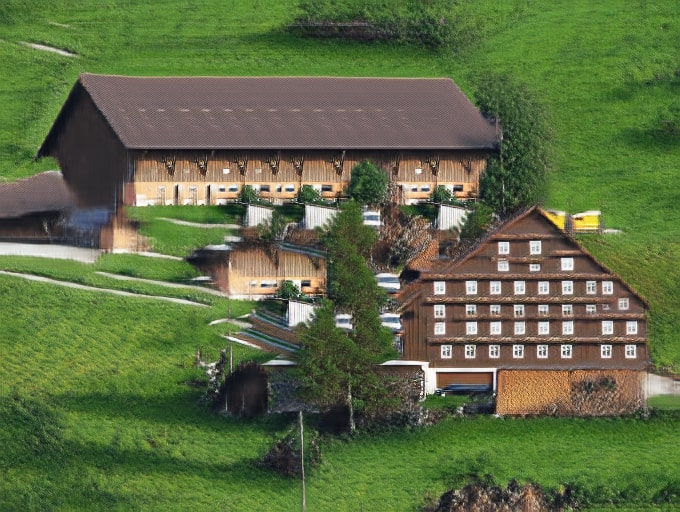} 

\end{tabular}
\caption{\textbf{Unified treatment for a range of datatypes}
\small \it
InGAN handles  textures, paintings and natural images with a single  architecture,
whereas texture synthesis methods~\cite{jetchev2016texture,nonstationary} poorly handle natural images, and retargeting methods~\cite{seamcarving-imp,simakov2008summarizing} struggle with textures.}
\label{fig:comparison_x2}
\vspace*{-0.5cm}
\end{figure*}

%% file: fig_comparison_textures.tex
\begin{figure*}
    \setlength\arrayrulewidth{3pt}
    \centering
    \begin{tabular}{c@{\hskip1pt}|@{\hskip1pt}c@{\hskip1pt}|@{\hskip1pt}c@{\hskip1pt}c@{\hskip1pt}|@{\hskip1pt}c@{\hskip1pt}c}
    input & Non-stationary \cite{nonstationary} &
    \multicolumn{2}{c@{\hskip1pt}|@{\hskip1pt}}{Spatial-GAN \cite{jetchev2016texture}} &
    \multicolumn{2}{c}{\textcolor{red}{\textbf{InGAN}}} \\
    
    \includegraphics[width=.1\linewidth, cframe=red 1.6pt]{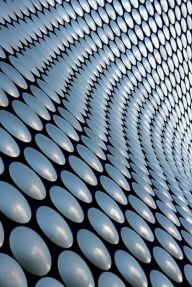} &
    \includegraphics[width=0.2\linewidth]{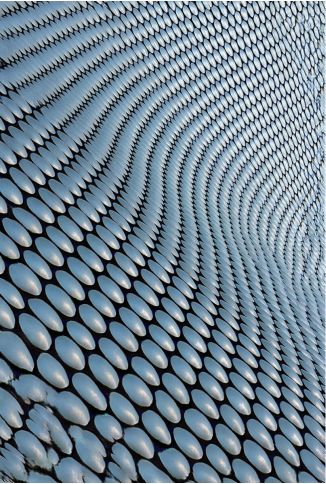} &
    \includegraphics[width=0.15\linewidth]{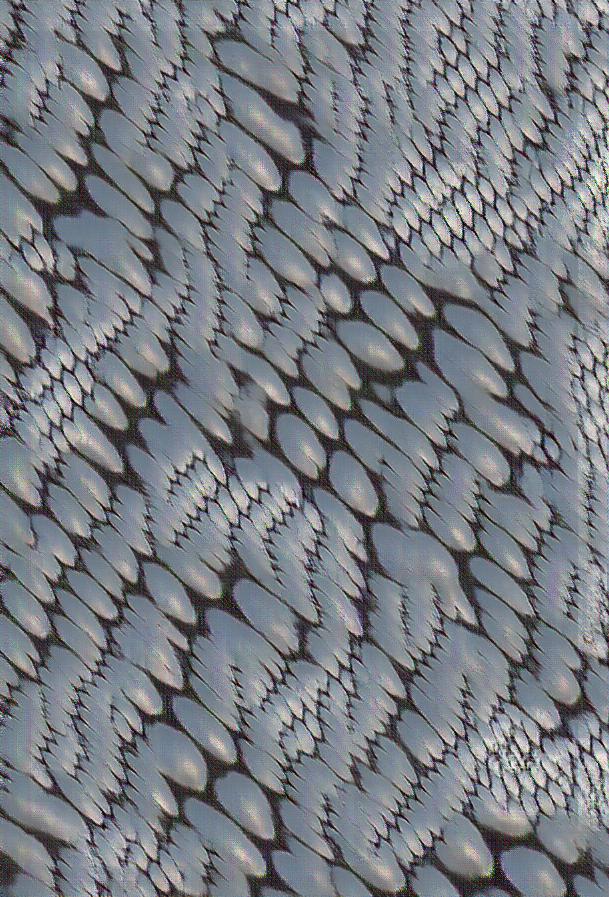}&
    \includegraphics[width=0.2\linewidth]{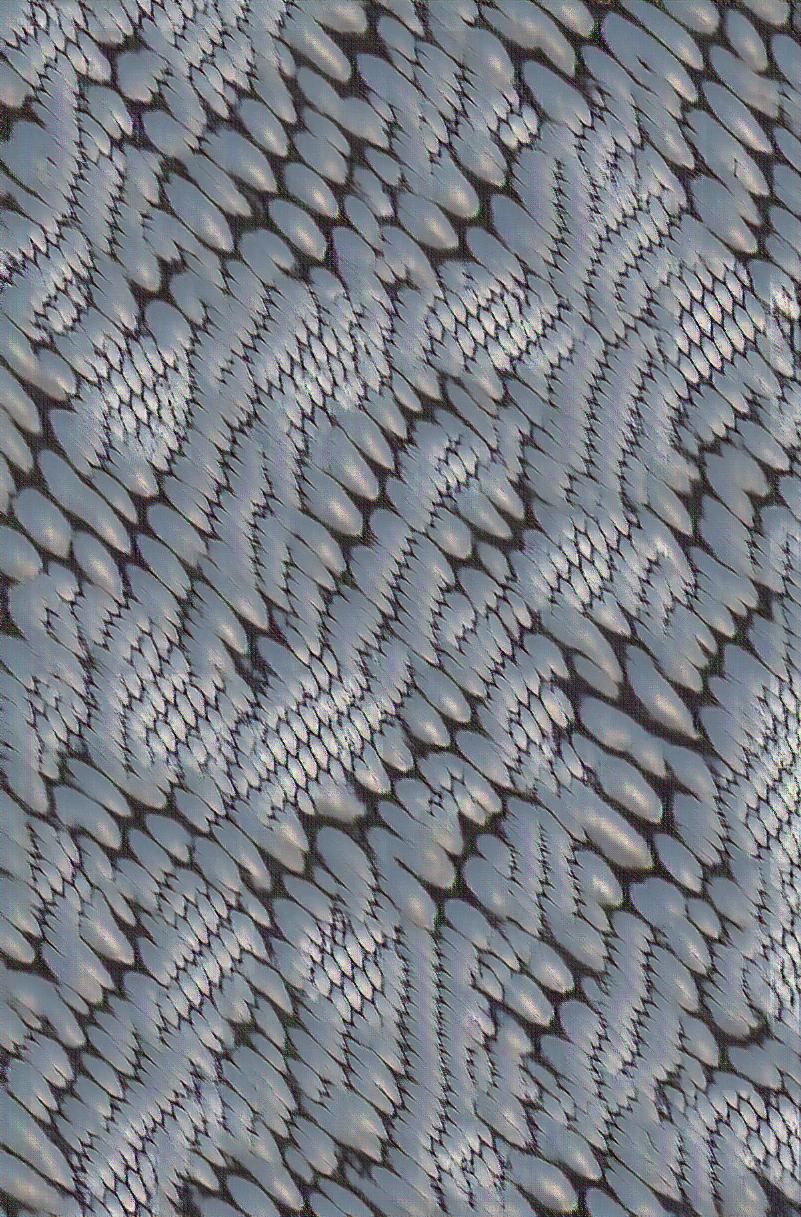} &
    \includegraphics[width=0.15\linewidth]{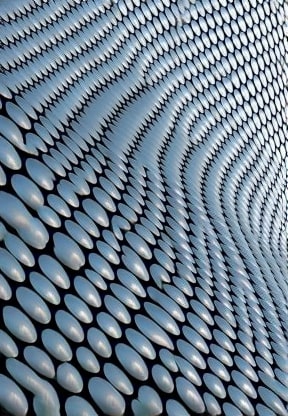} &
    \includegraphics[width=0.2\linewidth]{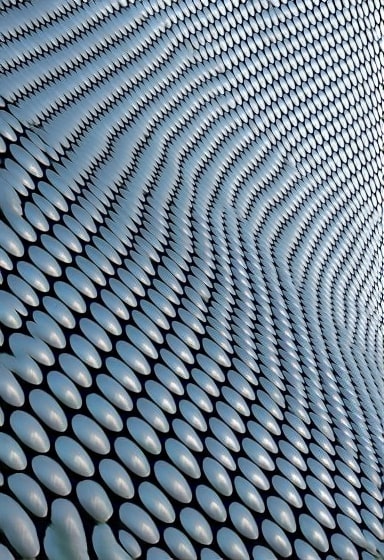} \\

    \includegraphics[width=.1\linewidth, cframe=red 1.6pt]{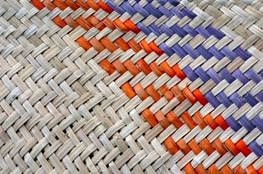} &
    \includegraphics[width=0.2\linewidth]{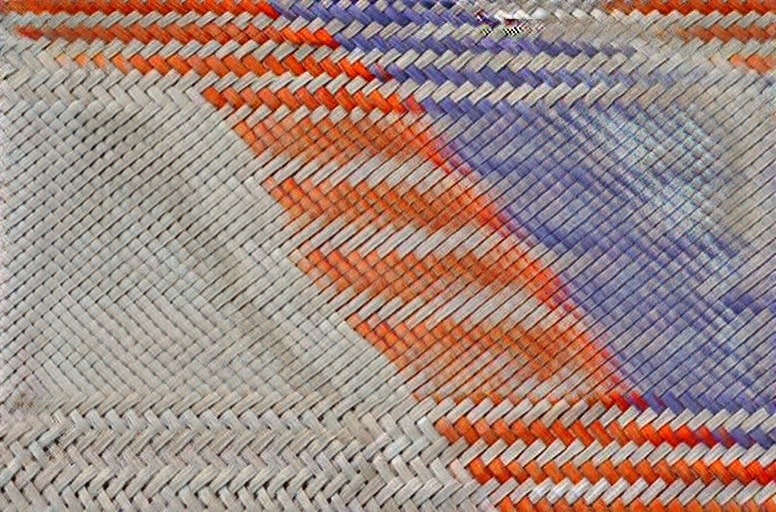} &
    \includegraphics[width=0.15\linewidth]{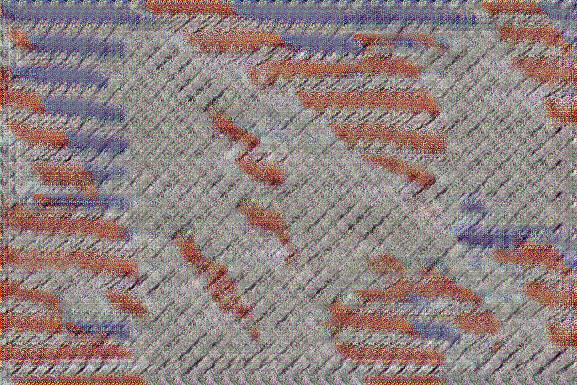} &
    \includegraphics[width=0.2\linewidth]{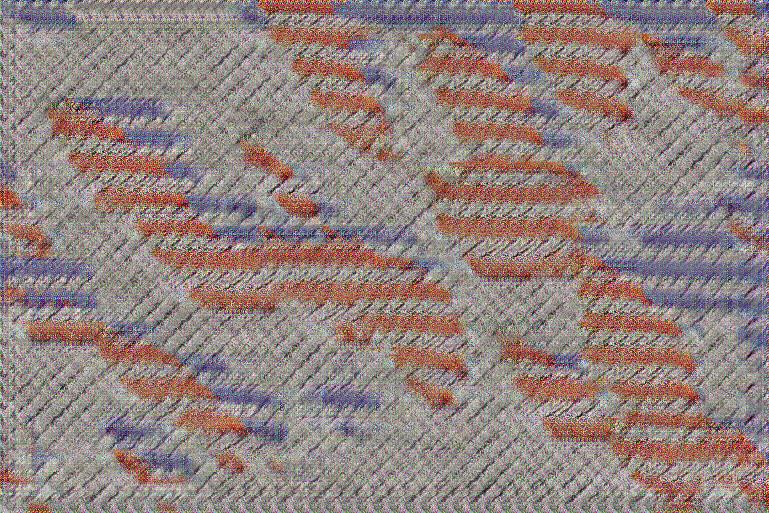} &
    \includegraphics[width=0.15\linewidth]{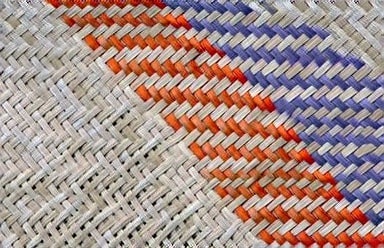} &
    \includegraphics[width=0.2\linewidth]{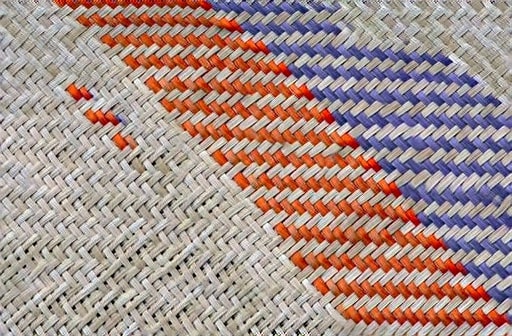} \\

    \includegraphics[width=.1\linewidth, cframe=red 1.6pt]{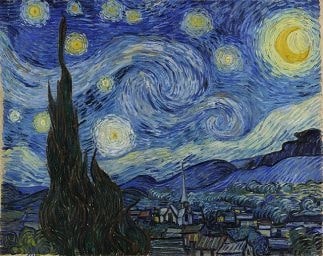} &
    \includegraphics[width=0.2\linewidth]{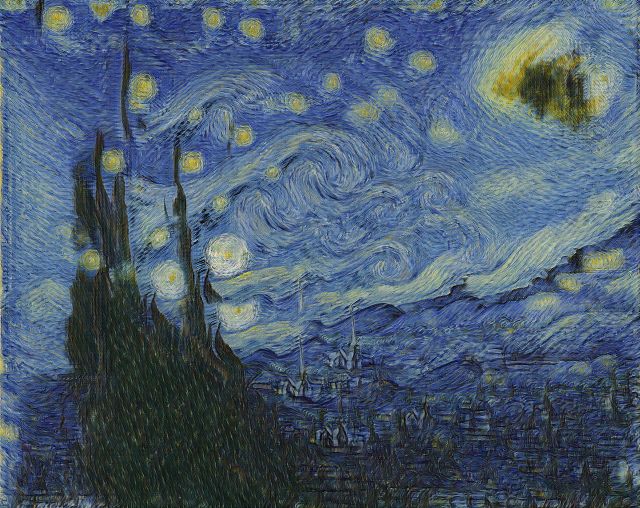} &
    \includegraphics[width=0.15\linewidth]{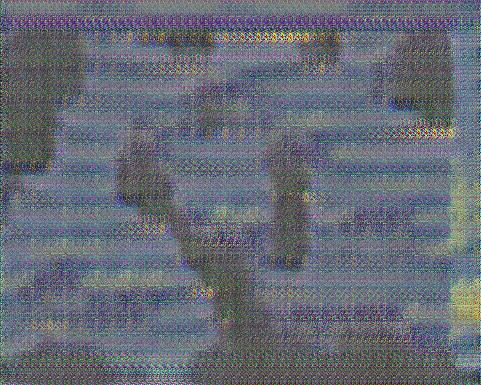} &
    \includegraphics[width=0.2\linewidth]{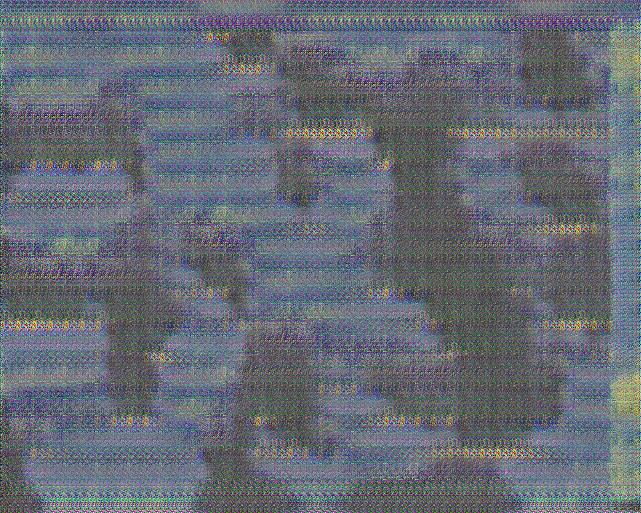} &
    \includegraphics[width=0.15\linewidth]{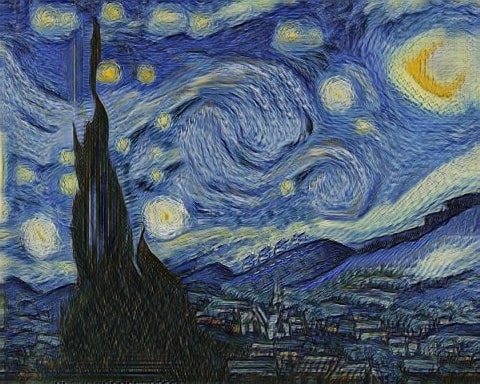} &
    \includegraphics[width=0.2\linewidth]{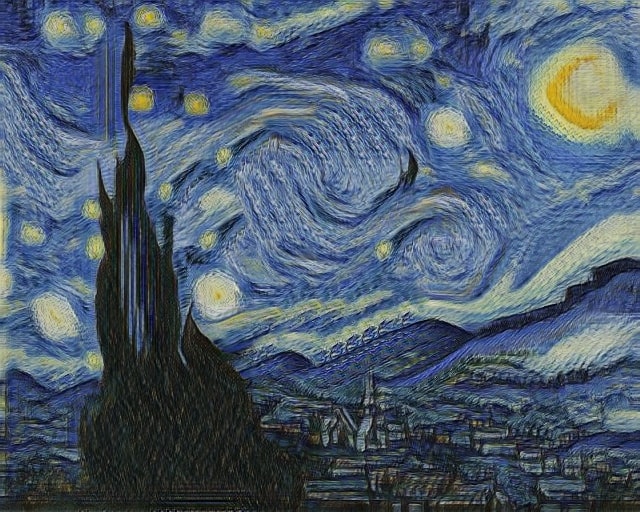} \\

    \includegraphics[width=.1\linewidth, cframe=red 1.6pt]{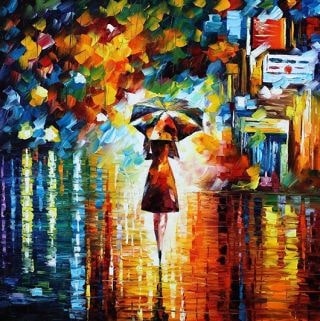} &
    \includegraphics[width=0.2\linewidth]{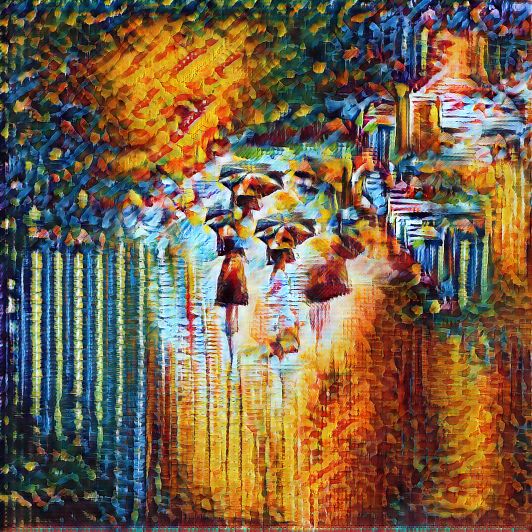} &
    \includegraphics[width=0.15\linewidth]{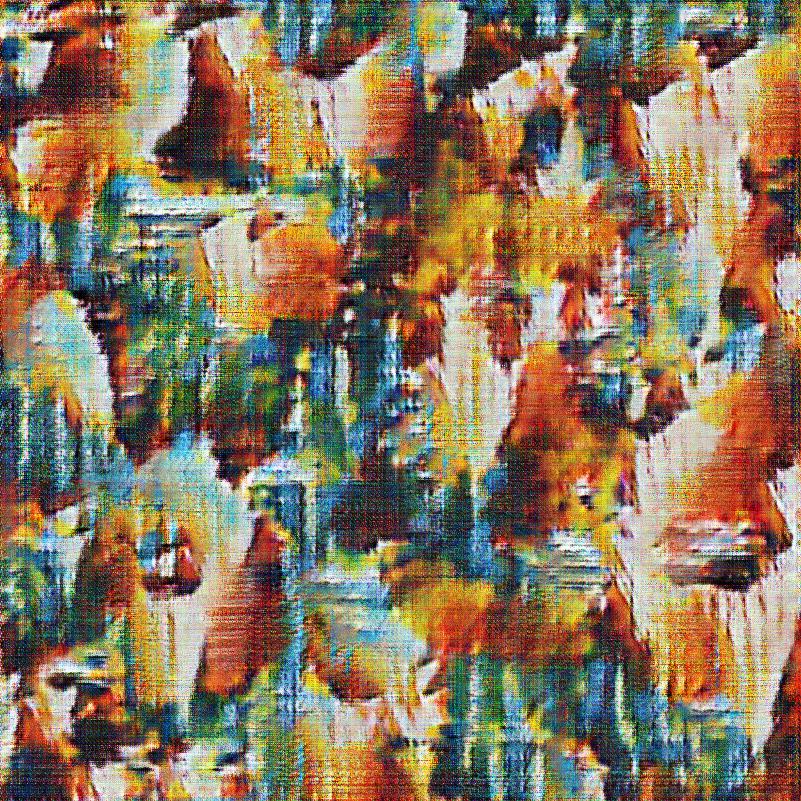} &
    \includegraphics[width=0.2\linewidth]{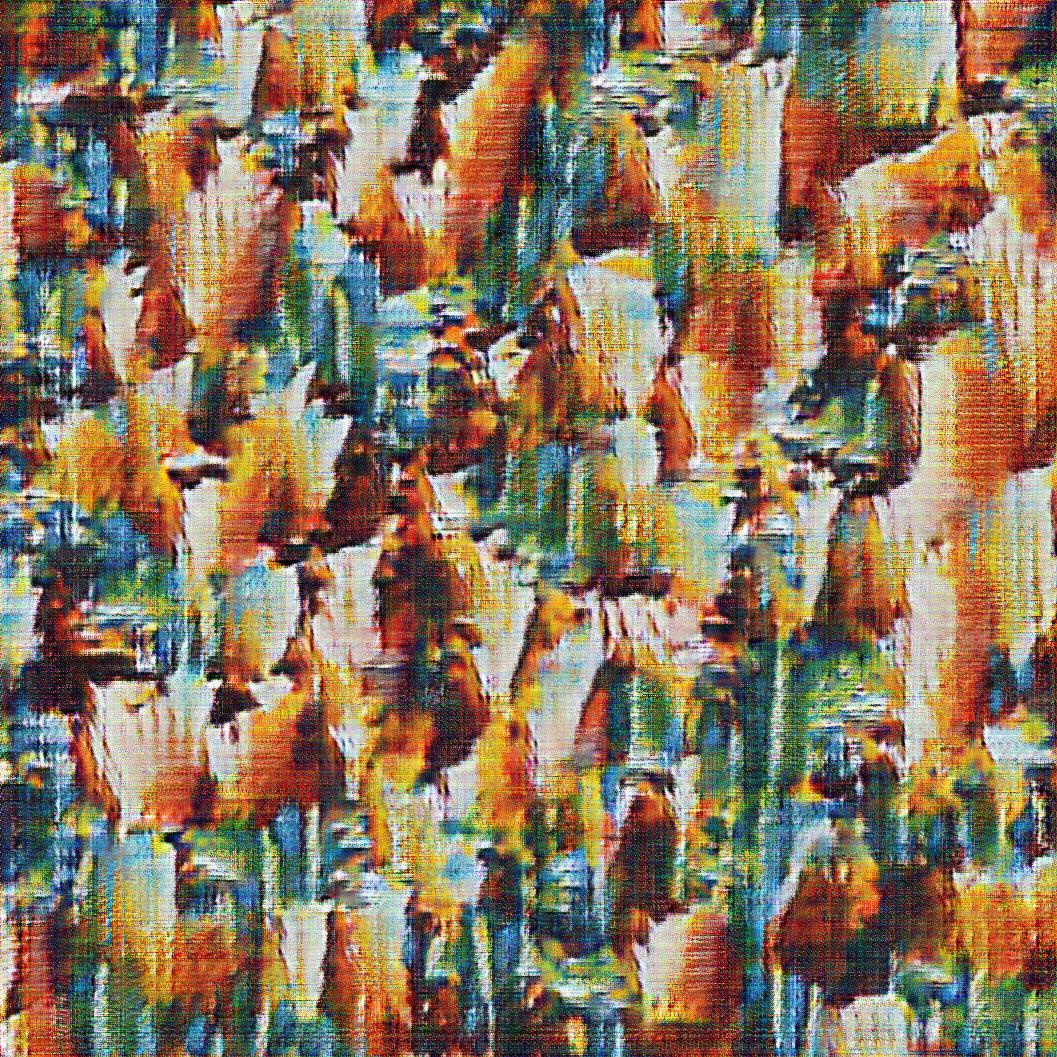} &
    \includegraphics[width=0.15\linewidth]{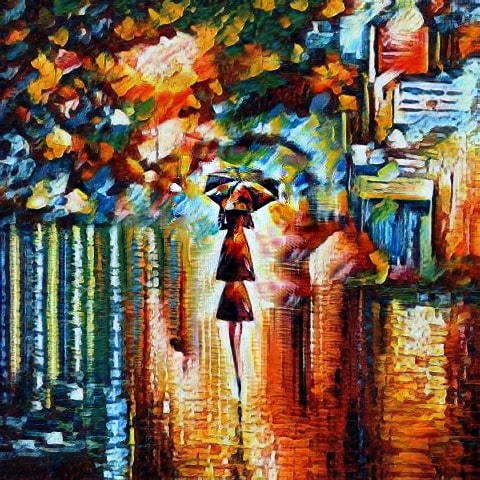} &
    \includegraphics[width=0.2\linewidth]{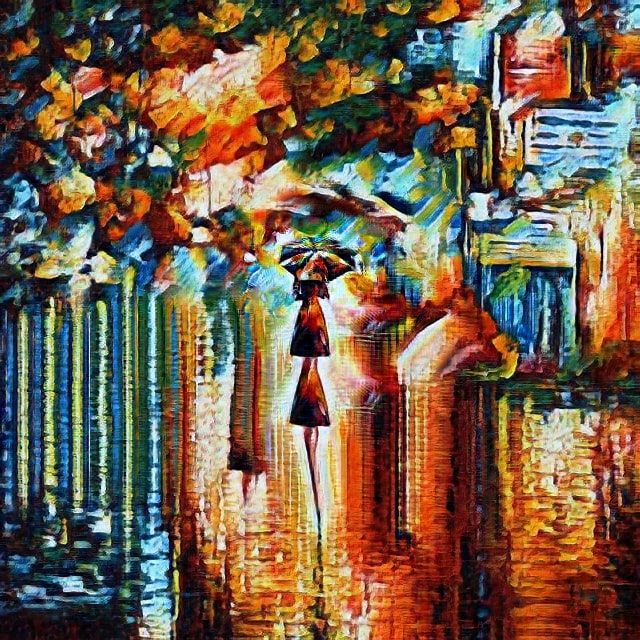} \\

    \end{tabular}
    \caption{\textbf{Texture synthesis:} 
    \small \it
    Synthesizing textures of sizes $\times 1.5$ and $\times 2$. Note that \cite{nonstationary} is restricted to  $\times2$. \emph{\textbf{\textcolor{red}{Please zoom in.}} }}
    \label{fig:comparison-textures}
    \vspace*{-0.3cm}
\end{figure*}

%% file: implementation.tex
\section{Implementation Details}
\label{optimization}
We use the ADAM optimizer \cite{adam} and a linearly decaying learning rate. 
We train over crops, ranging from $192\times192$ to $256\times256$, with a batch-size of 1. 
The default weighting of the $\Lrecon$ loss is $\lambda=0.1$. 
At each iteration, parameters of a Homography transformation $T$ are randomly sampled, resulting in different output size, shape and aspect ratio.
We employ a form of curriculum-learning so that the possible distortion allowed for $T$ is initially very small.
As the training progresses the allowed range of deformations gradually grows through the curriculum period (10k iterations) until it finally covers the entire desired range.
 
We employ several mechanisms for encouraging stability; spectral normalization~\cite{spectralnorm} is used both in the discriminator and the generator for all layers except the last one. Batch normalization~\cite{batchnorm} is used in most conv-blocks. 
We also encountered a degenerate case where $D$ was able to discriminate real patches from generated ones due to the fact that all values of the real patches were quantized to values $n/255$. To avoid this we add uniform noise in the range of $\left[0, 1/255\right]$ to the real examples before feeding them to the discriminator.

InGAN requries around 20k-75k iterations of gradient descent in order to obtain appealing results. Training takes 1.5-4 Hrs on a single V-100 GPU, regardless of the size of the input image. Once trained, InGAN can synthesize images of any desired size/shape/aspect-ratio in a single feedforward pass. 
For example, InGAN can remap to VGA size ($640$$\times$$480$) in about $40$ ms (equivalent to 25 fps).

%% file: fig_natural_image_remapping.tex
\begin{figure*}
    \centering
    \includegraphics[width=.8\linewidth]{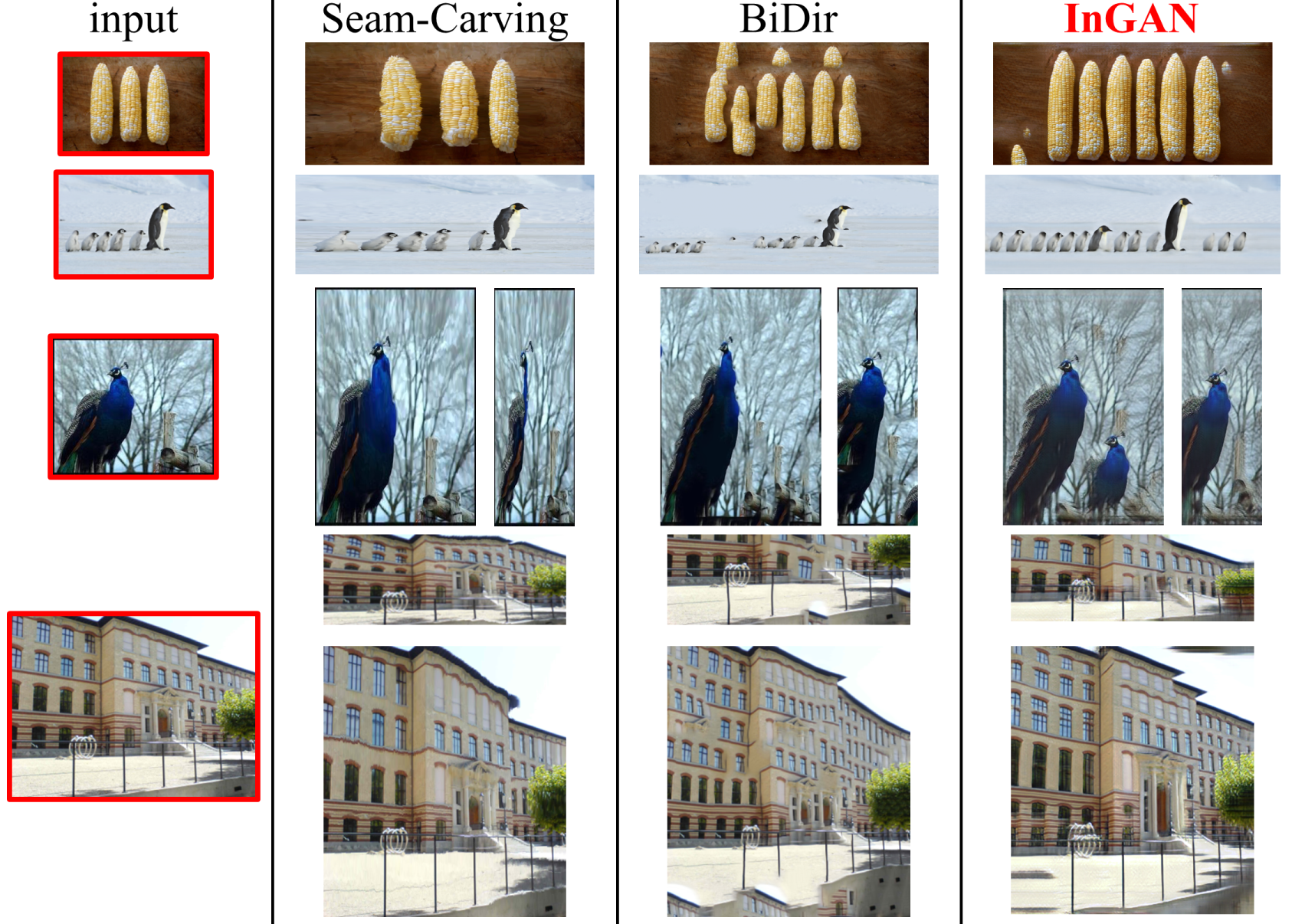}
    \caption{\textbf{Natural image retargeting:} 
    \small  \it 
    Comparing InGAN to Bidirectional similarity \cite{simakov2008summarizing} and Seam Carving \cite{seamcarving-imp}. \emph{\textbf{\textcolor{red}{Please zoom in.}} }}
    \label{fig:natural-image-remapping}
    \vspace*{-0.5cm}
\end{figure*}

%% file: unified-framework.tex
\section{A Unified Framework for Multiple Tasks}
\label{sec:unifiedFramework}
InGAN provides a variety of capabilities and can be applied to multiple tasks. Moreover, it provides a \emph{unified treatment of  very different data-types}, ranging from \emph{pure textures} to \emph{natural images}, all under a single umbrella.

A single pure texture is usually captured by just a few dominant image scales. Natural images, on the other hand, tend to span a wide range of image scales, from fine-grained details to coarse structures. Non-stationary textures and multi-textured images tend to reside somewhere in-between those two extremes. Each of these family of images is usually treated separately, by different specialized methods. Here we show that capturing and remapping the \underline{multiscale} patch distribution of an image provides a unified treatment for all these seemingly different data-types. InGAN thus applies to a \emph{a continuum from pure textures to natural images}, in a single framework.

For example, observe the corn image in Fig.~\ref{fig:natural-image-remapping}: small image patches at fine image scales capture the tiny details of the corn seeds, while patches at coarse images scales capture the structure of an entire corn-cob.
When retargeting the corn image to a wider/thinner output, \emph{entire corn-cobs are added/removed} (thus matching the multiscale patch distribution). In contrast, when changing the height of the output image, \emph{small corn seeds are added/removed} from each corn-cob.
This multiscale patch distribution is a fundamental characteristic of both natural images and textures.
Nonetheless it is important to stress that InGAN has no semantic information about ``objects'' or ``scenes'', it only models the multiscale patch distribution of the input image.

Figs.~\ref{fig:all_sizes},\ref{fig:comparison_x2},\ref{fig:comparison-textures},\ref{fig:natural-image-remapping},\ref{fig:homographies} exemplify the range of capabilities and data-types handled by InGAN. Additional examples are found in the Supp-Material. A unique capability of InGAN is its \emph{continuous transitions between different shapes/sizes/aspect-ratios}, \textbf{best exemplified by the attached videos in the Supplementary-Material}.

We next discuss a variety of tasks \& capabilities provided by InGAN,
all with a \emph{single network architecture}.  
InGAN may not provide state-of-the-art results
compared to \emph{specialized methods} optimized for a specific task (often also for a specific output size). Nevertheless, InGAN compares  favorably to these specialized methods, while providing a single unified framework for them all.
Moreover, InGAN  opens the door to \emph{new  applications/capabilities}.

\vspace*{0.15cm}
\noindent
\textbf{\underline{Texture Synthesis}:} \
Texture synthesis is the task of synthesizing a larger texture image from a small sample of the texture.
Gatys~\etal~\cite{gatys2015texture,style_transfer} used pretrained  network features to synthesize textures.
``Spatial-GAN''~\cite{jetchev2016texture} and ``Non-Stationary Texture Synthesis''~\cite{nonstationary} use a patch-based GAN in a fully convolutional manner, producing high quality textures. We refer to these kinds of textures (whether stationary or non-stationary) as \textbf{Single-texture synthesis}.
Texture synthesis methods typically perform poorly on  \textbf{Multi-texture synthesis} -- namely, synthesizing images containing multiple textures at various scales.
\mbox{InGAN} can handle both single- and multi-texture synthesis (as well as  natural images), thanks to its \emph{multiscale discriminator}.
Figs.~\ref{fig:comparison_x2} and~\ref{fig:comparison-textures}  
show comparisons of InGAN to specialized texture-synthesis methods, both on single- and multi-texture images (\cite{nonstationary} is restricted to $\times$$2$ outputs).

\vspace*{0.15cm}
\noindent
\textbf{\underline{Natural Image Retargeting: Summary and Expansion}} \
Image retargeting aims at displaying a \emph{natural} image on a different display size, smaller or larger, often with a different aspect ratio.
Smaller representations (visual summary, thumbnail) should faithfully represent the input visual appearance as best as possible. Another goal is to generate Expanded images of the \emph{same nature} (often with different aspect ratios).

There are several different notions of ``image retargeting''. Some methods (e.g.,~\cite{cho2017weakly,wolf2007}) aim at preserving salient objects while seamlessly discarding background regions to obtain a smaller image. They mostly do smart cropping, keeping the main object centered in its original size. Some of these methods struggle when there are several dominant objects present in the image. They do not tend to perform well on images with lots of clutter and texture, nor are they catered to image expansion/synthesis.
Seam-carving~\cite{seamcarving} gradually removes/adds pixel-wide ``seams" that yield  minimal change to image gradients. This method can handle both Summarization and Expansion. 

Other methods (e.g.,~\cite{simakov2008summarizing,pritch2009shift})  aim at preserving  local sizes/aspect-ratios of \emph{all}  image elements  (whether salient or not) as best possible, while changing the global size/aspect-ratio of the image. They cater both Summarization and Expansion. InGAN belongs to this family of methods.

Figs.~\ref{fig:all_sizes},\ref{fig:comparison_x2},\ref{fig:natural-image-remapping} show retargeting examples and comparisons of InGAN to Seam-Carving and Bidirectional-Similarity,
on natural images as well as non-natural ones.
Since Seam-carving \cite{seamcarving} uses local information only (it removes/adds pixel-wide ``seams"), 
it tends to distort larger image structures under drastic changes in aspect ratio (see narrow distorted peacock in Fig.~\ref{fig:natural-image-remapping}).
Bidirectional-Similarity~\cite{simakov2008summarizing} handles this  by using image patches at various scales, but
requires solving a new optimization problem for each  output size/aspect-ratio.
In contrast, InGAN  synthesizes a plethora of new target images of different sizes/aspect-ratios \emph{with a single trained network}. \textbf{Please view attached videos.}

\input{fig_homographies}

\vspace*{0.15cm}
\noindent
\textbf{\underline{Image Retargetig to \emph{Non-rectangular} Shapes}:} \
Unlike any previous method, InGAN is able to
retarget images into \emph{non-rectangular} outputs. This is made possible by introducing random \emph{invertible geometric transformations} in \mbox{InGAN's} Encoder-Encoder generator. Our current implementations uses Homgraphies (2D projective transformations), but the framework permits any invertible transformations.
Figs.~\ref{fig:all_sizes} and~\ref{fig:homographies} display a few such examples.
Note that a pure homography tilts all the elements inside the image. In contrast, InGAN preserves  local shape \& tilt of these elements despite  the severe change in global {shape} of the image. 
In particular, while the synthesized visual quality is not very high under {extreme} shape distortions,  InGAN generates an interesting illusion of retargeting into \emph{a new 3D view with correct parallax} (but without any 3D estimation).

%% file: fig_homographies.tex
\begin{figure*}
\vspace*{-0.6cm}
\centering
\includegraphics[width=1.8\columnwidth]{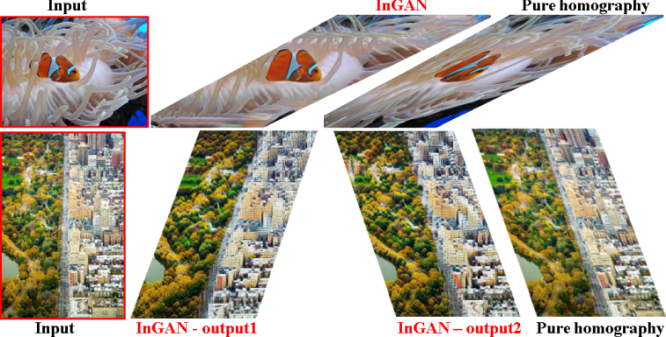}
\caption{\label{fig:homographies}
\textbf{Retargeting to Non-Rectangular Outputs:}
\small \it
InGAN is able to retarget to non-rectangular shapes using the geometric transforation $T$ (e.g., homography). Note that a pure homography tilts all the elements inside the image, wheras InGAN preserves  local shape/appearance \& tilt of these elements. In particular, InGAN generates an illusion of retargeting to a new 3D view with correct parallax (without any 3D recovery).
}
\vspace*{-0.6cm}
\end{figure*}

%% file: experiments.tex
\section{Ablation Study and Limitations}
\label{sec:ablation}
\label{sec:failure_cases}

\input{fig_ablation}

We conducted an ablation study to verify the importance of: (i)~the ``encoder-encoder" architecture with its $\ell_1$ reconstruction loss, and (ii)~the importance of multiple scales in the discriminator $D$. 
Fig.~\ref{fig:ablation} shows one such example:
Training InGAN without $\Lrecon$ (left-most result) shows unstructured output: two birds are completely missing and the dominant bird is split into two.
Using a single scale $D$ (middle result) makes $G$ generate a result that is locally coherent, but lacks large scale structures. The birds were completely destroyed.
In contrast, the full InGAN (right-most) with $\Lrecon$ and multiscale $D$ maintains both fine details and coarse structures. In particular, all 3 birds are in the  output.

\input{fig_fail}
\vspace*{0.15cm}
\noindent \textbf{Limitations:}
InGAN is trained in an unsupervised manner -- it has no additional information other than the input image itself. 
In particular, InGAN has no semantic understanding, no notion of ``objects" or ``scenes". Its sole objective is capturing and remapping the multiscale patch distribution of the input image.
Thus, InGAN sometimes produces funny/unnatural results. 
Fig~\ref{fig:fail} shows such an example: 
\mbox{InGAN} produces an output that is both \emph{coherent} and \emph{complete} (all local elements are preserved), yet is incorrect in its semantic meaning.

\comment{

During our experiments we noticed two types of failure cases. Fig.~\ref{fig:fail} shows an example of each type: 

\vspace*{0.15cm}
\noindent
\todo{replace failure examples here?} 

\vspace*{0.15cm}
\noindent \textbf{Local Orientation:}
InGAN requires all patches of the output image to match the distribution of patches in the input. Transformations/deformations over these patches can be generalized, but only if sufficient diversity in the input patches exist.
When the input image contains certain elements oriented only in one direction, e.g., the contrails of the airplanes in Fig.~\ref{fig:fail}, the learned distribution cannot generalized to more possible orientations, leading to severe artifacts in the retargeted image.
}
\comment{
Note that the last statement does not mean that the output patches have to be identical to the input patches. The distribution of patches is subject to the interpretation of the discriminator. It may grant high probability to patches that do not appear in the input image at all, as long as they are interpreted as drawn from the learned distribution.
\\
This limitation is apparent in cases where the input image contains certain elements oriented only in one direction. In such cases the learned distribution is not generalized to more possible directions. These cases are the internal equivalent of having a training-set that does not contain a sufficient variety of examples.
In most cases, we are not bothered by this limitation as we usually prefer elements to preserve their orientation, but there are cases where we would want patches to change their aspect ratio. Typical such cases are those where the image contains long lines. Observe Fig.\ref{fig:fail} right hand side; The retargeted image is indeed comprised from patches drawn from the patches distribution of the original input image. Each patch along the trail holds a line with an incline that matches the incline of the trail in the original input image. However this is not the desired result.} 

\comment{
\vspace*{0.15cm}
\noindent \textbf{Object partition ambiguity:}
InGAN is trained in an unsupervised manner on a single image, and thus lacks semantic understanding. As a result, we observe that objects are sometimes split in two. 
This phenomenon is visible in the soldiers example: when retargeting to a larger image height, InGAN stops treating eac soldier as a single ``object" and duplicates their legs.
}

\comment{{{

\section{Experimental Results}
We used InGAN to capture and remap the internal statistics of a variety of images. Figures \ref{fig:all_sizes}, \ref{fig:comparison_x2}, \ref{fig:capitol}, \ref{fig:single-texture}, \ref{fig:comparison_scale_w} and \ref{fig:expansion} show some qualitative results. We encourage the reader to view additional results in the supplementary material.

\todo{Leave this subsection - but make it very concise}

\subsection{Analysis of results}
\noindent \textbf{Capturing multiscale patch distribution (the ``DNA" of an image):} 
Unlike single-texture images, a natural image exhibits both fine grained details as well as coarse level structures. For example, look at the corn image in Fig.~\ref{fig:all_sizes}: small patches capture the minute details of the corn seed while coarse patches capture the structure of an entire corn-cob. 
InGAN is able to capture this multiscale patch distribution: when retargeting the corn image into a thinner/wider output entire corn-cobs are removed or added. Also when changing the image height, corn seeds are added or removed from each corn-cob. However, at no point neither the corn cobs nor the corn seeds are distorted by InGAN despite the radical change in aspect ratio between input and output.
This multiscale patch distribution is a fundamental characteristic of natural images.
InGAN's ability to capture it well can be observed also in other examples. 
For instance look at the fruit stand of Fig.~\ref{fig:comparison_x2} and the birds of Fig.~\ref{fig:comparison_scale_w}. 
A good visualization of this capability of InGAN is shown in the video clips submitted in the supplementary material and we encourage the reader to view them.

\comment{ 
For example, imagine you are given an input image $I$, and you wish to transform it to a new image $J$, of drastically different shape, size and aspect ratio. But you don't want to distort any of its internal elements; you want to keep them all in their original size, shape, aspect ratio, and in the same relative position within the image. Such examples are shown in Figs.~\ref{fig:farmhouse,fig:fruitstand}. Note that despite changing the global size and shape of the farmhouse image, the windows in the target images maintain their local size and shape. Rows of windows are automatically added/removed, and likewise for the number of windows in each row. Similarly, when the fruit-stand image is enlarged (Fig.~\ref{fig:fruitstand}), more fruits are added in each fruit-box while keeping the size of each fruit the same; and vice versa – when the image grows smaller, the number of fruits grows smaller, while maintaining their size and their relative position within the image.
}

\noindent \textbf{Element generalization:} 
Looking at the rice field image of Fig.~\ref{fig:capitol} (left), it depicts three adults and one child. 
When retargeting the image to twice its height (top most result) InGAN adds one adult  but also generates one young person, bigger than the smallest child but smaller than the adults. 
InGAN is able to generalize new elements from the learned distribution.
Note that InGAN does not duplicate elements. Having more than one example, it can generalize and synthesize novel instances. 
%
Nonetheless it is important to stress that InGAN has no semantic information about ``objects" or ``scenes", it only models the multiscale patch distribution of the input image.

\comment{
\noindent \textbf{Local distribution matching:} \label{sec:local_dist}
Fig.~\ref{fig:results} demonstrates the principal of local distribution matching; 
When retargeting menu items (top) or soldiers (bottom) InGAN matches the local distribution of elements in the images. New sushi dishes are synthesized at the relative location of sushi dishes, while new nigiri dishes are synthesized at their respective location. The same goes for the marching soldiers: InGAN synthesizes new soldiers with the appropriate uniform color relative to their local distribution.
This local distribution matching property of InGAN gives rise to stability of retargeting results as the target size is being gradually changed. This stability is best observed in video clips attached in supplamentary material.
}
\comment{
\subsection{Comparison to other methods}
Fig.~\ref{fig:comparison_x2} visually compares InGAN to bidirectional similarity (BiDir)~\cite{simakov2008summarizing}, seam carving (SC)~\cite{seamcarving} and non-stationary texture synthesis~\cite{nonstationary}\footnote{We were unable to compare to~\cite{cho2017weakly}: code or models were not made publicly available. Moreover, there is no project webpage or any other reliable source for the input images used in their study.} on the challenging task of retargeting images to twice their input size.
BiDir exhibits very high quality locally, however, it struggles with maintaining intact global structures in the image, e.g., the roof of the farm house is bent, the crates of the fruit stand are shuffled.
SC makes very local decision per added seam, thus when applied for such extreme retargeting it completely distorts the image.
The method of \cite{nonstationary}, despite its similarity to InGAN, is only capable of $\times 2$ enlarging of a \emph{single} texture. When applied to natural images with many, multiscale textures it fails to produce visually pleasing results.
In contrast to these methods, InGAN is capable of preserving both locality and multiscale textures and produce visually pleasing results even in these extreme retargeting settings.

Fig.~\ref{fig:comparison_scale_w} shows further comparisons to BiDir and SC on the task of retargeting to smaller size than input size (note that we cannot compare to~\cite{nonstationary} in these settings as it is restricted to $\times 2$ retargeting). 
BiDir consistently produces very pleasing results locally, but struggles with very global structures: it crops the tips of the carrots and completely remove the dominant bird.
SC distorts the output images in these retargeting scenarios.
Again, InGAN is able to produce visually pleasing results by showing an entire carrot as well as preserving all three birds.

Additional comparison results as well as video clips showing comparisons of gradual retargeting of images can be found in the supplementary material.
}
\comment{
Although at first glance the method of \cite{nonstationary} seams similar to InGAN: training GAN on a single image, it is 

We ran several comparison experiments for comparing to other methods. \cite{nonstationary, simakov2008summarizing, seamcarving} 
We find that results by \cite{simakov2008summarizing} consist of valid patches, which due to overlapping creates consistent images. However, since optimizing general requirement for bidirectional similarity over the whole image, the locality is only implicit and tends not to hold, especially when upscaling. Examples can be seen in Fig.~\ref{fig:comparison_x2} Where the fruit types are mixed and not locally matching and also the farm that lacks reasonable structure. Fig.~\ref{fig:comparison_scale_w} where only two birds are kept and another branch appears at the top of the image. Distribution matching is a stronger criterion than Bidirectional similarity. Fig.~\ref{fig:summarizing} where the result of \cite{simakov2008summarizing} is close to being a crop from the input image rather than a visual summary that maintains the distribution of patches as the InGAN result.
While Bidirectional similarity optimization takes several to several tens of minutes, one optimization process is only valid for one scale of the image.
\\
Seam Carving \cite{seamcarving} achieves satisfying results for images that contain several main objects. However, we find in our experiments that it lacks the ability to handle textures and tends to distort the image. Examples for such distortion can be seen in Fig.~\ref{fig:comparison_x2} and in Fig.~\ref{fig:comparison_scale_w}. The vulnerability to textures is apparent in Fig.~\ref{fig:summarizing}. The reason for this poor result is the fact the seams of low derivatives are carved, this keeps only the high derivatives pixels resulting in less smooth textures.
\\
Zhou~\etal~\cite{nonstationary} shows impressive result for non-local textures, but we find that it cannot handle textures of varying scales nor does it have the capacity for the multi-texture synthesis needed for natural images. Both findings are apparent in the fruit image at the bottom of Fig.~\ref{fig:comparison_x2}. Notice that while able to synthesize watermelons texture, the smaller fruits get blurred. The farm image in the same figure demonstrates the lacking of capacity for multiple textures and preseving non-texture objects.
Zhou \etal is trained for relatively long time on a single texture and is then only applicable for upscaling in a factor of 2.
\\
}

\subsection{Ablation Study} \label{sec:ablation}
\input{fig_ablation}
Fig.~\ref{fig:ablation} shows an example of ablation study we conducted to verify the importance of (1)~the ``encoder-encoder" architecture and the $\ell_1$ reconstruction loss, and (2)~the importance of multiscales in the discriminator $D$. We used the birds input image (Fig.~\ref{fig:comparison_scale_w}). 
Training InGAN without $\mathcal{L}_\texttt{reconstruct}$ (left most result) shows unstructured output: two birds are completely missing and the dominant bird is split into two.
Using single scale $D$ (middle result) makes $G$ generate a result that is locally coherent, but lacks large scale structures. The birds were completely destroyed.
In contrast, full InGAN (right most result) with $\mathcal{L}_\texttt{reconstruct}$ and multiscale $D$ maintains both fine details and coarse structures in the retargeted output.

\subsection{Limitations and Failure Cases}
\label{sec:failure_cases} 
\todo{replace examples here}

During our experiments we noticed two types of failure cases. Fig.~\ref{fig:fail} shows an example of each type:
\\
\noindent \textbf{Local Orientation:}
InGAN requires all patches of the output image to match the distribution of patches in the input. Transformations/deformations over these patches can be generalized, but only if sufficient diversity in the input patches exist.
When the input image contains certain elements oriented only in one direction, e.g., the contrails of the airplanes, the learned distribution cannot generalized to more possible orientations leading to severe artifacts in the retargeted image. 
\comment{

Note that the last statement does not mean that the output patches have to be identical to the input patches. The distribution of patches is subject to the interpretation of the discriminator. It may grant high probability to patches that do not appear in the input image at all, as long as they are interpreted as drawn from the learned distribution.
\\
This limitation is apparent in cases where the input image contains certain elements oriented only in one direction. In such cases the learned distribution is not generalized to more possible directions. These cases are the internal equivalent of having a training-set that does not contain a sufficient variety of examples.
In most cases, we are not bothered by this limitation as we usually prefer elements to preserve their orientation, but there are cases where we would want patches to change their aspect ratio. Typical such cases are those where the image contains long lines. Observe Fig.\ref{fig:fail} right hand side; The retargeted image is indeed comprised from patches drawn from the patches distribution of the original input image. Each patch along the trail holds a line with an incline that matches the incline of the trail in the original input image. However this is not the desired result.}
\\
\noindent \textbf{Object partition ambiguity:}
InGAN is unsupervisingly trained on a single image thus lacks semantic understanding. This fact prevents modeling of what is a full object that can only exist as a complete unit and what can be partitioned into standalone sub-objects. This phenomenon is visible in the soldiers example: when retargeting to larger image height, InGAN stops treating the soldiers as a single ``unit" and duplicates their legs only.
\comment{
The multi-scale discriminator compromises between scales and prefers images with patches that optimally satisfy all the scales. When upscaling, this mechanism needs to determine what scale of element will get duplicated at each location. In order to do so it needs information regarding what is a valid stand-alone object. In some cases, such information can be obtained from the image statistics, but in other cases it is lacking and can cause non-realistic output. In the left hand side of Fig.\ref{fig:fail}, the generated output duplicated the soldiers' legs. In this particular case, the small patches in the connecting line between the two lines of legs are reasonable since the upper area of the original line of legs also contains dark diagonal stripes. The localization attribute of InGAN preserves the local distribution and thus prefers adding more legs. A similar phenomenon is apparent for the heads of the soldiers, but in this case the human observer finds it more reasonable as it may look like these heads belong to another line of soldiers. This difference of our evaluation of the image is semantic and therefore cannot be used by InGAN.  }
}}}

%% file: fig_ablation.tex
\begin{figure}
\centering
\begin{tabular}{c@{\hskip2pt}c@{\hskip3pt}c@{\hskip3pt}c}
\includegraphics[width=0.48\columnwidth, cframe=red 1.6pt]{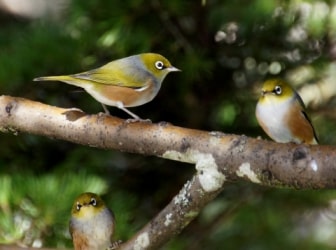} &
\includegraphics[width=0.16\columnwidth]{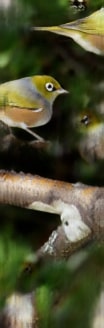} &
\includegraphics[width=0.16\columnwidth]{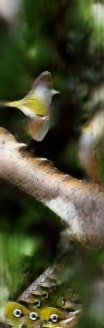} &
\includegraphics[width=0.16\columnwidth]{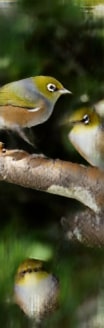} \\
Input & 
No & 
Single- &
\textbf{\textcolor{red}{InGAN}} \vspace*{-0.1cm}\\
 & 
$\Lrecon$ & 
scale $D$ &
 \\
\end{tabular}
    \caption{\textbf{Ablation study:}
    \small \it
    Omitting $\Lrecon$ or using a single-scale $D$, degrades the results compared to  full \textcolor{red}{InGAN} architecture.}
    \label{fig:ablation}
    \vspace*{-0.75cm}
\end{figure}

%% file: fig_fail.tex
\begin{figure}
\centering
\includegraphics[width=\columnwidth]{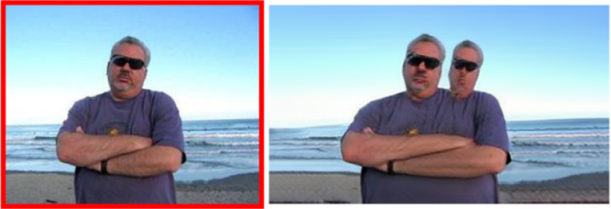}
\caption{\label{fig:fail}
\textbf{Failure example:}
\small \it
Input in \textcolor{red}{red}. 
\small \it
InGAN  has  no semantic understanding of ``objects'' or ``scenes'',  it only models the multiscale patch distribution of the input image, hence cannot distinguish between object-parts and entire objects.
}
\vspace*{-0.35cm}
\end{figure}

%% file: acknowledge.tex
\noindent\textbf{Acknowledgements:} This project has received
funding from the European Research Council (ERC) under
the European Union’s Horizon 2020 research and innovation
programme (grant agreement No 788535). Additionally, supported by a research grant from the Carolito Stiftung. Dr Bagon is a Robin Chemers Neustein Artificial Intelligence Fellow.